\documentclass{article}

\usepackage{microtype}
\usepackage{graphicx}
\usepackage{subcaption}
\usepackage{booktabs} 

\usepackage{makecell}
\usepackage{bbding}

\usepackage{graphicx}
\usepackage{subcaption}
\usepackage{float}
\usepackage{caption}
\usepackage{lscape}                                         
\usepackage{wrapfig}

\usepackage{animate} 

\usepackage{booktabs}                   
\usepackage{multirow}
\usepackage{arydshln} %

\usepackage{enumitem}

\usepackage{bm}                          

\usepackage{amssymb}
\usepackage{amsmath}  

\usepackage{units}
\usepackage{color}
\usepackage[T1]{fontenc}    
\usepackage{amsfonts}       
\usepackage[utf8]{inputenc} 
\usepackage{nicefrac}       
\usepackage{microtype}      
\usepackage{pifont}         
\usepackage{comment}

\usepackage[mathscr]{euscript}
\usepackage{colortbl}

\usepackage{url}  

\def\vs{\emph{vs.}~}                 

\newlength\paramargin

\newcommand{\mfigure}[2]
{
\includegraphics[width=#1\linewidth]{#2}
}

\newcommand{\secref}[1]{Section~\ref{sec:#1}}
\newcommand{\figref}[1]{Figure~\ref{fig:#1}} 
\newcommand{\tabref}[1]{Table~\ref{tab:#1}}

\long\def\ignorethis#1{}

\newbox\jsavebox%

\graphicspath{{figure}, {example}}

\usepackage{xcolor}
\colorlet{dark-blue}{blue!50!black}
\colorlet{dark-cyan}{cyan!75!black}
\colorlet{dark-purple}{purple!50!black}
\colorlet{dark-red}{red!75!black}
\colorlet{dark-green}{green!75!black}
\colorlet{dark-orange}{orange!50!black}
\colorlet{dark-gray}{black!75}
\colorlet{light-gray}{black!30}
\definecolor{nice-red}{HTML}{E41A1C}
\definecolor{nice-orange}{HTML}{FF7F00}
\definecolor{nice-yellow}{HTML}{FFC020}
\definecolor{nice-green}{HTML}{39b54a}
\definecolor{nice-blue}{HTML}{0071bc}
\definecolor{nice-purple}{HTML}{984EA3}

\colorlet{verylight-gray}{black!10}
\definecolor{LightCyan}{rgb}{0.88,1,1}

\definecolor{best}{rgb}{1, 0.85, 0.7}
\definecolor{second}{rgb}{1,1, 0.8}

\graphicspath{{figures/}}

\usepackage{xspace}
\newcommand{\etal}{\textit{et al.}\xspace}
\newcommand{\eg}{e.g.\xspace}

\usepackage{hyperref}

\usepackage[capitalize,noabbrev]{cleveref}

\usepackage[preprint]{icml}

\usepackage{amsmath}
\usepackage{amssymb}
\usepackage{mathtools}
\usepackage{amsthm}

\theoremstyle{plain}

\theoremstyle{definition}

\theoremstyle{remark}

\makeatletter
\renewcommand{\ICML@preprint}{}
\makeatother

\icmltitlerunning{HaRP: High Dynamic Range Photosequencing}

\begin{document}

\twocolumn[
  \icmltitle{HaRP: High Dynamic Range Photosequencing through \\ Dual Reversed Shutter Scanning}



  \icmlsetsymbol{equal}{*}

  \begin{icmlauthorlist}
    \icmlauthor{Xiang Ji}{yyy}
    \icmlauthor{Guixu Lin}{yyy}
    \icmlauthor{Jiancheng Zhao}{yyy}
    \icmlauthor{Zhengwei Yin}{yyy}
    \icmlauthor{Yinqaing zheng}{yyy}
  \end{icmlauthorlist}

  \icmlaffiliation{yyy}{Graduate School of Information Science and Technology, the University of Tokyo, Tokyo, Japan}

  \icmlcorrespondingauthor{Xiang Ji}{ji-xiang@ynl.t.u-tokyo.ac.jp}
  \icmlcorrespondingauthor{Yinqiang Zheng}{yqzheng@ai.u-tokyo.ac.jp}

  \icmlkeywords{Machine Learning}

  \vskip 0.3in
]



\printAffiliationsAndNotice{}  

\begin{abstract}
The adoption of CMOS sensors in mobile photography is frequently compromised by the rolling shutter (RS) effect, which introduces geometric distortions and motion artifacts. Particularly, recent rolling shutter with global reset (RSGR) mode, while mitigating some RS issues, also incurs major limitations, including reduced capture speed and compressed dynamic range. To address these problems, we propose a novel dual reversed scanning setup utilizing both RSGR and inverted RSGR views. This solution not only handles the inherent flaws of RSGR by synchronizing complementary exposures to balance the dynamic range across the frames but also introduces an effective method for HDR photosequencing under highly dynamic scenes. Our proposed network first accommodates row-wise complementarity and manages visual shifts by row-adaptive feature alignment. Subsequently, the hallucination module, built upon a correlation-guided mix-attention block, integrates the mutually reinforced features to recover missing details. In addition, we construct a coaxial imaging system to collect a real-world dataset, enabling robust training and evaluation beyond numerical simulation. Experimental results demonstrate the twofold benefits of our solution in mitigating RSGR limitations and advancing HDR reconstruction techniques.
\end{abstract}

\section{Introduction}
\label{sec:intro}

With the widespread use of digital cameras and smartphones, increasing attention has been paid to choosing imaging sensors that balance performance and cost. Compared with Charge-Coupled Devices (CCD), Complementary Metal-Oxide Semiconductor (CMOS) sensors dominate the consumer market due to their low power consumption, high data rate, and ease of integration~\cite{vasu2018occlusion,schubert2019rolling}, and have become fundamental to visual computing~\cite{kirillov2023segment,ji2018dense,ye2020drm,zhao2025tree}. However, under dynamic objects or ego-motion, their row-wise scanning mechanism produces rolling shutter (RS) effects, degrading both appearance and geometry. As a result, mitigating RS effects has become an essential problem for robust scene understanding and reconstruction.

\begin{figure*}[!tb]
	\centering
	\mfigure{0.96}{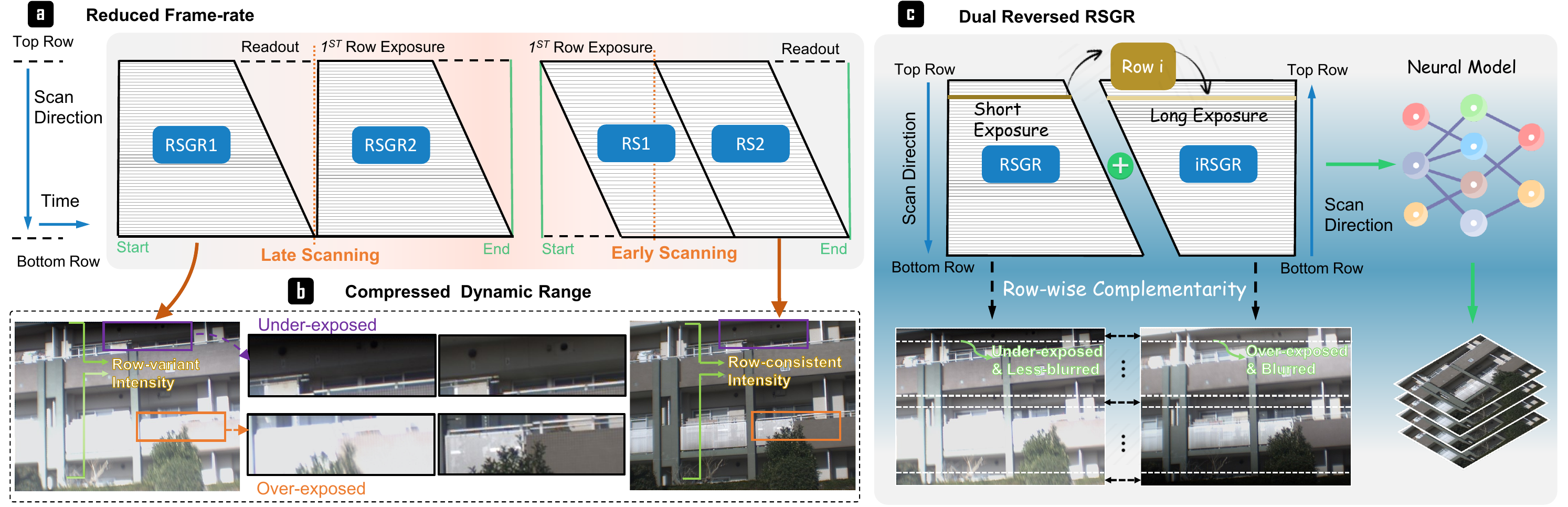}
	\vspace{-1mm}
	\caption{
		Limitations of RSGR mode and our dual reversed setting. (a) shows capturing process of two consecutive frames from RSGR and RS sensors. Enabling the global reset feature deprives RS mode of early scanning and causes the delay of a readout time. (b) presents acquired frames from two types of sensors. The dynamic range of RSGR is lower than that of RS. In (c), we reconstruct an HDR video clip from two row-wise complementary observations captured by spatiotemporally aligned RSGR (scanning downwards) and iRSGR (scanning upwards) cameras, which well addresses two issues of RSGR mode and provides a novel solution for HDR reconstruction.
	}
	\label{fig:motivation}
	\vspace{-4mm}
\end{figure*}

Traditional RS correction methods~\cite{albl2020two,zhuang2017rolling,lao2018robust,purkait2017rolling,rengarajan2016bows} rely on velocity assumptions or geometric constraints, which struggle to capture real camera motion and scene geometry. Deep learning-based approaches have achieved impressive performance for single~\cite{zhuang2019learning,yan2023deep} and multi-frame inputs~\cite{liu2020deep,ringaby2012efficient,fan2023joint,qu2023towards,niu2024rs}. Recent works~\cite{fan2021inverting,fan2022context,zhong2022bringing} further attempt to recover multiple latent frames from RS inputs, but the ill-posed nature of the problem still leads to ghosting and motion artifacts. To handle this issue, the global-reset feature of RS sensors (RSGR)~\cite{wang2022neural,ji2023single,ji2023rethinking} has been exploited to reformulate RS correction as a more tractable deblurring problem, achieving superior results. But meanwhile, the RSGR mode introduces two overlooked limitations: (1) the merit of RS mode over global shutter (GS) in capture speed is sacrificed because early scanning of the next frame is not allowed anymore (\figref{motivation}a); and (2) row-variant exposure causes dynamic-range compression (\figref{motivation}b). Unlike conventional RS with constant row exposure, RSGR increases exposure duration row by row, leading to under-exposure at the top and over-exposure at the bottom when scanning downward.

Existing HDR approaches cannot be directly applied, because they depend on alternating-exposure LDR sequences combined with motion compensation~\cite{kang2003high,chen2021hdr} or feature alignment ~\cite{chung2023lan,liu2022ghost}. Such methods remain vulnerable to distortions and ghosting under fast motion, making them ineffective for handling the limitations. Inspired by ~\cite{albl2020two}, we propose a dual reversed setting simultaneously capturing two fully aligned RSGR and inverted RSGR (iRSGR) views to reconstruct a sequence of clear frames with high speed and high dynamic range (HDR photosequencing), which not only well-addresses two aforementioned issues but also provides a novel solution for HDR reconstruction in dynamic scenarios. As shown in~\figref{motivation}c, the corresponding rows in the two views provide complementary exposures for the same visual content under static conditions. Although the middle scanlines may converge to similar intensity levels, missing contrast can be compensated by the hallucination capability of our model. In dynamic scenes, visual shifts between corresponding rows may weaken this complementarity. However, such shifts are typically minor due to the short exposure time, and our method remains robust to them. Meanwhile, these shifts also alleviate intensity convergence in the middle rows. Our experiments in~\secref{visual_shift} confirm these conclusions. Since this row-wise complementarity holds in both static and dynamic scenes, our dual reversed RSGR setup ultimately enables high-frame-rate acquisition, mitigates motion degradation, and makes dynamic-range expansion more tractable.

Following the hardware design of ~\cite{zhong2022bringing,ji2023rethinking,ji2024motion,ji2026moment}, we construct our optical system to capture aligned high-speed sharp HDR videos and low-speed RSGR-iRSGR pairs. Assisted by the collected dataset, we further propose an end-to-end framework for HDR photosequencing. Firstly, the Row-adaptive Feature Alignment module (RFA) conducts multi-scale, row-shaped attention to better adapt the row-wise complementarity and simultaneously manage the visual shifts under highly dynamic scenes. Subsequently, the hallucination module generates sharp details, especially for under/over-exposed rows. To this end, we introduce a residual correlation-guided mix-attention block (RCMB), built upon parallel streams, which aggregates intra-stream self-similar features and supports cross-stream mutual compensation. Finally, a tri-pivot temporal loss on the luminance channel emphasizes geometric consistency and suppresses color variations, improving temporal coherence in the reconstructed video. 
In summary, the main contributions of our paper are:
\vspace{-4mm}
\begin{itemize}[leftmargin=*]
	\item We design a dual-reversed RSGR configuration that inherently accommodates fast motions, addressing intrinsic limitations of RSGR mode, and simultaneously offering a competitive solution to robustify HDR reconstruction.
	\vspace{-3mm}
	
	\item We construct a special imaging system that simultaneously captures RSGR-iRSGR pairs along with their corresponding high-speed HDR videos, and collect a high-quality real dataset of outdoor dynamic scenes. 
    \vspace{-3mm}
	
	\item We introduce a novel neural network architecture consisting of a row-adaptive feature alignment and a hallucination module based on correlation-guided mix-attention block. Extensive experiments have validated the effectiveness of our setting and model.
	
\end{itemize}

\section{Related Work}
\label{sec:relate}

 \begin{figure*}[!tb]
	\centering
	\mfigure{0.95}{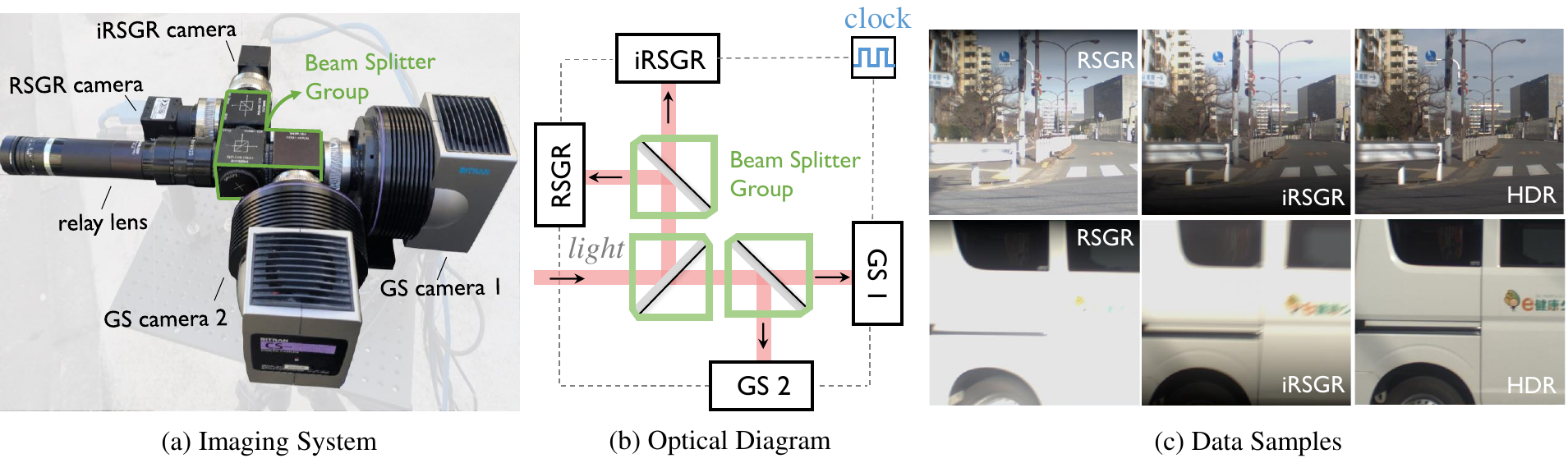}
	\vspace{-2mm}
	\caption{
		(a) shows the constructed data capturing platform that is spatially aligned and temporally synchronized through a beam splitter group and a signal controller, respectively. (b) is the corresponding optical diagram. We also present some data samples in (c).
	}
	\label{fig:imaging_system}
	\vspace{-6mm}
\end{figure*}
\subsection{Rolling Shutter Correction and Interpolation}

The causing effects of rolling shutters under ego-motion or dynamic scene are often non-negligible. Early works model RS geometry~\cite{meingast2005geometric} to improve SfM and correct distortions using discrete epipolar constraints~\cite{dai2016rolling} or differentiable warping~\cite{zhuang2017rolling}, but they rely on strong assumptions about scene structure or camera motion, such as Manhattan-world layouts~\cite{purkait2017rolling}, straight-line preservation~\cite{rengarajan2016bows}, purely rotational~\cite{lao2018robust,ringaby2012efficient}  or translational motion~\cite{liang2008analysis,baker2010removing}. These priors limit their applicability in real scenes. Albl \etal~\cite{albl2020two} later introduce a dual reversed RS configuration that provides additional geometric constraints from sparse correspondences.

Learning-based approaches~\cite{naor2022combining,zhou2022evunroll,fan2021sunet,zhuang2019learning} instead estimate motion fields to warp RS images toward their GS counterparts. Liu \etal~\cite{liu2020deep} propose a shutter-unrolling network with differentiable warping, while Qu \etal~\cite{qu2023towards} solve complex nonlinear motion with a quadratic RS model. Fan \etal~\cite{fan2023joint} jointly recover appearance and motion in a single-stage design, and the global-reset mode has also been exploited for video~\cite{wang2022neural} and single-image correction~\cite{ji2023single}. More broadly, RS interpolation or temporal super-resolution~\cite{fan2021inverting,fan2022context,zhong2022bringing} reconstructs multiple latent sharp frames within a single RS exposure, and Ji \etal~\cite{ji2023rethinking} provide a unified comparison across degradations. Nevertheless, RS correction fundamentally requires shifting each scanline to a virtual GS coordinate frame based on estimated flow or hallucinated context, making existing methods fragile in the presence of large or complex motions.

\subsection{High Dynamic Range Imaging}

Regardless of HDR video reconstruction or multi-exposure imaging, the goal is to merge motion-varying LDR frames into a clean HDR result. Traditional HDR pipelines focus on alignment to avoid ghosting: motion rejection methods~\cite{khan2006ghost,heo2010ghost,zhang2011gradient,oh2014robust,lee2014ghost} first register LDR inputs and then discard misaligned pixels, often suppressing useful information and degrading HDR quality. More robust techniques use optical flow~\cite{kang2003high,zimmer2011freehand} or energy minimization~\cite{sen2012robust,hu2013hdr}, but they remain time-consuming and sub-optimal due to complex optimization.

Deep learning has significantly advanced HDR imaging. CNN- and Transformer-based methods~\cite{kalantari2017deep,cai2018learning,wu2018deep,yan2019attention,yan2020deep,niu2021hdr} alleviate misalignment and reduce ghosting, even enabling single-image HDR reconstruction~\cite{chen2023learning,marnerides2018expandnet,liu2020single}. Kalantari \etal~\cite{kalantari2017deep} pioneer optical-flow-guided HDR fusion; Wu \etal~\cite{wu2018deep} avoid explicit motion estimation via image translation. Chen \etal~\cite{chen2021hdr} develop a coarse-to-fine alignment and fusion framework using their collected real dynamic HDR data. More recent works include context-aware Transformers~\cite{liu2022ghost} and luminance-based alignment with hallucination modules~\cite{chung2023lan} for enhanced detail recovery.

Some methods employ event cameras~\cite{lin2023event,zou2021learning} to reconstruct HDR videos, but such systems are often costly or vulnerable to large motions. In contrast, our dual reversed RSGR configuration is simple to implement and provides improved robustness under dynamic scenes.

\begin{figure*}[!tb]
	\centering
	\mfigure{0.9}{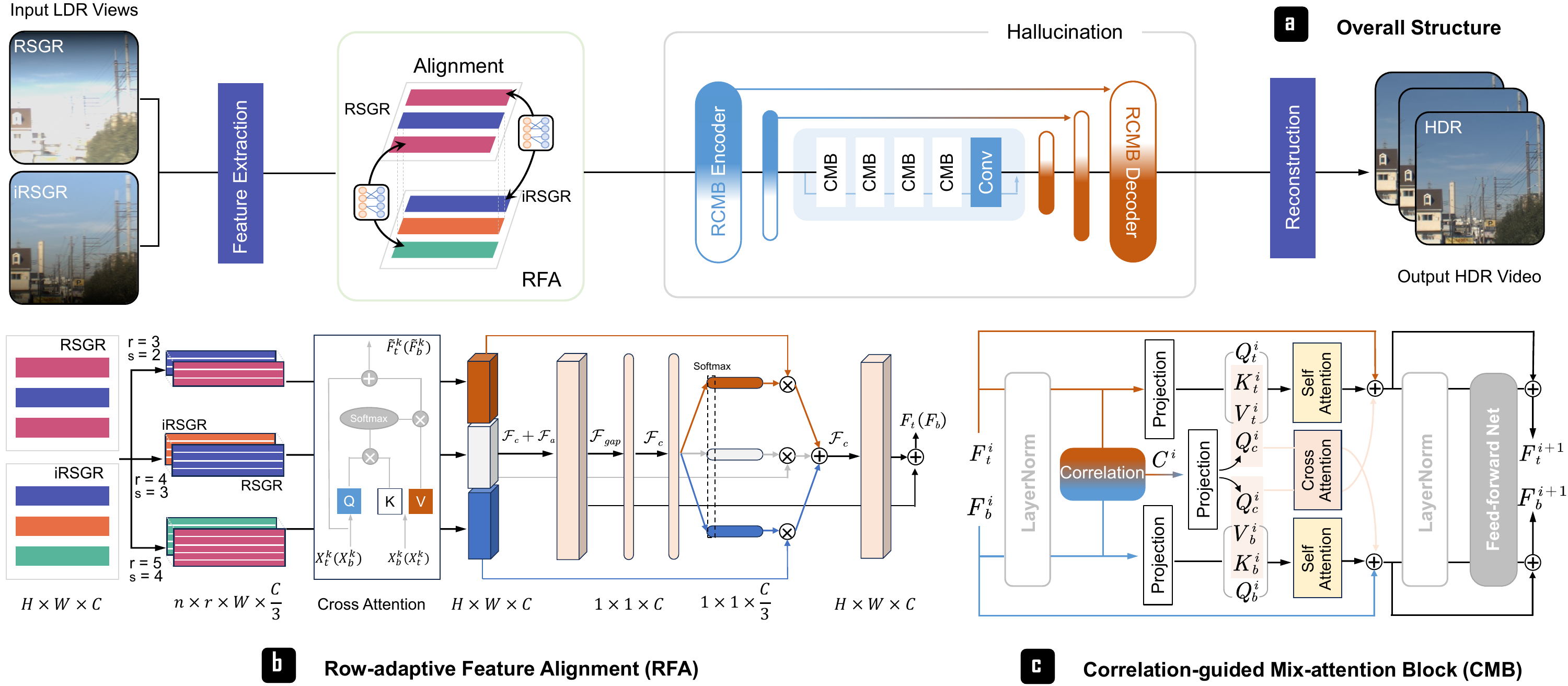}
	\vspace{-2mm}
	\caption{
		(a) shows the overall architecture containing two parts: alignment module registers motions between reversed RSGR views, and hallucination module generates sharp details, especially for under/over-exposed rows. Instead of simply associating corresponding rows between views, we exploit row-adaptive alignment strategy with multiple scales to handle the visual shift under highly dynamic scenes as shown in (b). (c) presents our correlation-guided mix-attention block that further facilitates the generation of local high frequencies. 
	}
	\label{fig:model}
	\vspace{-6mm}
\end{figure*}

\begin{figure}[!tb]
	\centering
	\mfigure{1}{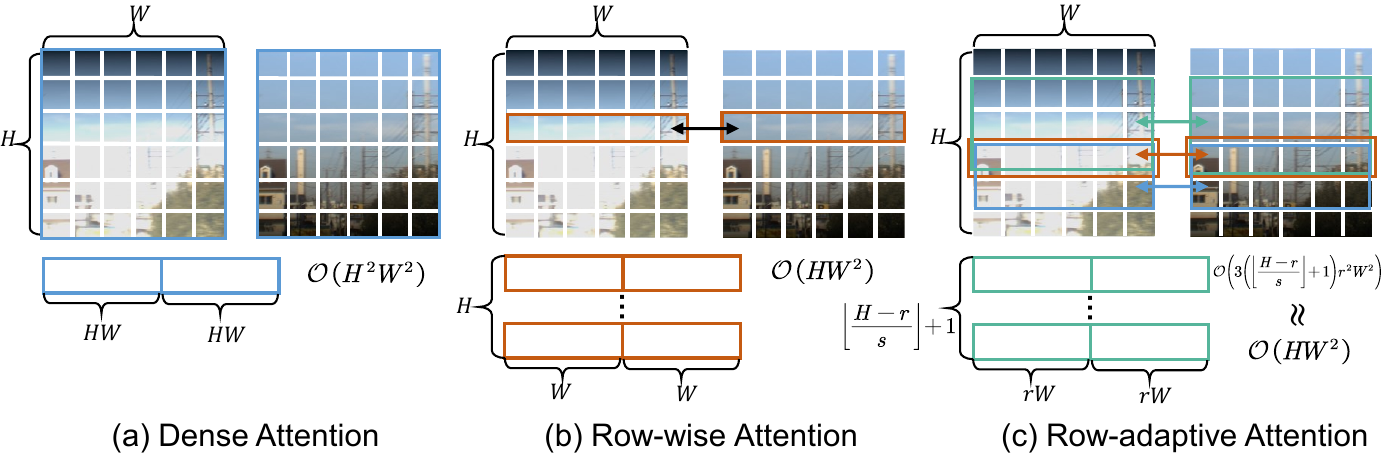}
	\vspace{-4mm}
	\caption{
		Different alignment strategies based on attention mechanism. We compared the full image, row-wise with single scale and row-adaptive with multiple scales alignment methods. Row size $r$ and stride $s$ are constants.
	}
	\label{fig:alignment}
	\vspace{-4mm}
\end{figure}

\section{Methodology}

\subsection{Optical System and RSGR-HDR Dataset}
\label{sec:data_capture}
To train our model, we require spatially aligned and temporally synchronized RSGR-iRSGR observations together with corresponding high-frame-rate HDR ground truths. Therefore, we develop a novel quad-axis imaging system and collect a real-world dataset, RSGR-HDR. Each capture provides a dual reversed RSGR pair and two sets of high-speed GS frames for composing HDR GTs. To the best of our knowledge, such real-world data have not been explored in prior work. Our entire
hardware prototype is illustrated in~\figref{imaging_system}. Two RSGR cameras (FLIR BFS-U3-63S4C) with reversed scanning direction share the identical objective scenes with two high-speed GS cameras (BITRAN CS-700C) through a beam splitter group. The RSGR sensors enlarge pitch size from $2.4~\mu m$ to $9.6~\mu m$ through $4 \times 4$ binning, while GS cameras incorporate SONY IMX426 sensor with air-forced cooling. We follow the merging strategy in~\cite{lin2023event,zou2021learning} to combine the GS pairs with different scene irradiance into HDR measurements.

Consequently, the RSGR-HDR dataset is constructed using our optical system, which consists of three modes: RSGR, inverted RSGR (iRSGR) and corresponding HDR frames, acquired under highly dynamic urban scenes. In total, we collect 90 videos with resolution of $640 \times 480$, each of which contains $70 \times 2$ degraded frames, accompanied by 1,400 aligned high-speed HDR frames. Notably, the characteristic differences of reversed RSGR views are quite obvious (\figref{imaging_system}c). More details about imaging system and dataset are provided in~\secref{system_details} and ~\secref{data_training}.

\subsection{Network Architecture}

To perform HDR photosequencing from dual reversed RSGR views, we propose a two-stage framework comprising feature alignment and hallucination modules, as shown in~\figref{model}a. Given strictly aligned RSGR–iRSGR pair, shallow convolutions first extract their feature maps $X_t$ and $X_b$. Row-shaped attention is then applied at multiple scales to handle visual shifts under large motion and to promote mutual compensation between the two views. Features aligned with row sizes $\bm{r}$ and strides $\bm{s}$ are dynamically fused to accommodate varying motion magnitudes. The hallucination module further aggregates these aligned features using a Transformer-based encoder–decoder to recover fine details in under- or over-exposed regions. Finally, a reconstruction block outputs a high-speed HDR sequence $\bm{H}=\{H^i, i= 0,1,\cdots,N-1\}$ within the exposure duration of input. N is the length of reconstructed latent video.  

\paragraph{Row-adaptive Feature Alignment}
Normally, attention-based alignment will attend to all tokens without selection or adhering to epipolar constraints. However, to effectively adapt and fully leverage our dual reversed RSGR configuration, several considerations need to be addressed. As discussed in~\cite{ji2023single,wang2022neural}, the unique distortion induced by RSGR mode is naturally row-dependent. Indiscriminate processing tends to corrupt the temporal cues implicitly embedded by sequential readout mechanism. Second, the observed RSGR and iRSGR views are inherently row-wise corresponding, facilitating robust alignment even under large motions. Furthermore, applying dense attention across the entire feature space (~\figref{alignment}a) leads to high computational demands and potential ambiguities in motion registration. Therefore, we propose row-shaped attention to highlight these concerns. Nevertheless, direct use of row-wise attention with fixed scale (\figref{alignment}b), \eg simply associating corresponding rows between views~\cite{shi2021stereo,li2024era3d} is only suitable for ideal static scenes. To handle the varying visual shifts under dynamic cases, we propose row-adaptive feature alignments (~\figref{alignment}c) that effectively blend multiple scales. For clarity, we proceed to explain with $\bm{r}=[3,4,5]$ and $\bm{s}=[2,3,4]$.

We first compute the bilateral alignments for dual views on multiple scales as shown in ~\figref{model}b and then dynamically fuse them in a learnable manner. By taking the RSGR branch as an example, the extracted features of RSGR and iRSGR views are denoted as $X_t, X_b \in \mathbb{R}^{H\times W\times C}$. On each scale level $k$, the features are accordingly converted to $X_t^k, X_b^k \in \mathbb{R}^{n \times r^k \times W \times C/3}$ by following row size $r^k$ and stride $s^k$, where $n = \lfloor  \frac{H-r^k}{s^k}+1\rfloor$. We generate the query $Q_t^k$ for $X_t^k$, and the key $K_b^k$, value $V_b^k$ for $X_b^k$ as follows:
\begin{equation}
	\small
	\begin{split}
		Q_t^k,K_b^k,V_b^k = X_t^k \ast K_1, X_b^k \ast K_2, X_b^k \ast K_3,
	\end{split}
	\vspace{-1mm}
\end{equation}
where $K_1,K_2,K_3$ denote $1\times 1$ kernels and $\ast$ is convolutional operator. Thus, the aligned feature under corresponding scale $\tilde{F}_t^k$ is formulated as:
\vspace{-2mm}
\begin{equation}
	\small
	\begin{split}
		\tilde{F}_t^k = X_t^k + (\sigma(Q_t^k(K_b^k)^T)V_b^k)\ast K_4.
	\end{split}
	\vspace{-2mm}
\end{equation}
$\ast K_4$ represents $1\times 1$ kernel convolution to match the number of channels with that of $X_t^k$. $\sigma(\cdot)$ is the softmax function.

The aligned features under all scales naturally retain motions with different magnitudes, which can be dynamically fused to make robust alignments. In more detail, we use the concatenated information of all scales to produce respective weights for each branch, then integrate them via weighting. Specifically, we first compute the fused feature $\tilde{F}_t^{fuse}$ using a fully connected layer $\mathcal{F}_{c}$ and an activation layer $\mathcal{F}_{a}$ as:
\vspace{-2mm}
\begin{equation}
	\begin{split}
	\tilde{F}_t^{fuse} = \mathcal{F}_a(\mathcal{F}_c(\oplus_{k=1,2,3}\tilde{F}_t^{k})),
	\vspace{-1mm}
	\end{split}
\end{equation}
where $\oplus$ means the concatenation of feature maps along the channel dimension. Subsequently, the weights are formulated as:
\vspace{-4mm}
\begin{equation}
	\small
	\begin{split}
		\tilde{F}_t^{w} &= \mathcal{F}_c(\mathcal{F}_{gap}(\tilde{F}_t^{fuse})) \\
		\alpha_k &= \frac{e^{\delta_k(\tilde{F}_t^{w})}}{\sum_{j=1}^{3}e^{\delta_j(\tilde{F}_t^{w})}}, k=1,2,3,
		\vspace{-1mm}
	\end{split}
\end{equation}
where the $\mathcal{F}_{gap}$ denotes global average pooling and $\delta_k$ is an operation to uniformly partition channel dimension  according to scale levels and return the $k^{th}$ part. Finally, the aligned feature for RSGR branch is computed by:
\begin{equation}
	\small
	\begin{split}
		F_t = \tilde{F}_t^{fuse} + \mathcal{F}_c(\sum_{k=1}^{3} \alpha_k \times \tilde{F}_t^k)
		\vspace{-1mm}
	\end{split}
\end{equation}
The generation of aligned feature $F_b$ for iRSGR branch is similar to the aforementioned process.

\paragraph{Correlation-guided Mix-attention Block}
To enhance the generation ability of local details, we further propose novel correlation-guided mix-attention Transformer layer with dual streams. Therefore, the aggregated features of both views can be mutually improved more effectively from perspective of global context. The calculated correlation volume~\cite{sun2018pwc} between two views explicitly represents the similarities, which has merits of driving the features in each stream to become more compatible with those of the other. To this end, we build the cross-attention module aided by the correlation volume as the major supplementary to original self-attention. Overall, our proposed mix-attention effectively adapts to self-similar features within a single stream while also enabling reciprocal interactions across different streams.

As shown in ~\figref{model}c, we present the specific operations for the $i^{th}$ CMB layer. Given input features $F_t^i$ and $F_b^i$ after layer normalization,
we compute their correlation volumes $C^i$ following the method in ~\cite{sun2018pwc,fan2021sunet} by setting the search range as $3\times 3$ to balance between complexity and accuracy. Note that, to ensure input order invariance, we consider the bidirectional nature of cost volumes and project concatenation of them into the same dimension as input features. For the self-attention, $F_t^i$ and $F_b^i$ are linearly projected to obtain their corresponding queries, keys and values: $\{Q_t^i, K_t^i, V_t^i\}$ and $\{Q_b^i,K_b^i,V_b^i\}$. The correlation volume $C^i$ is also projected as a query $Q_c^i$ serving as guidance for cross-attention. Formally, we define the computational process as:
\begin{equation}
	\small
	\begin{split}
		F_{t\_agg}^i = \sigma(\frac{Q_t^i(K_t^i)^T}{\sqrt{d_k}})V_t^i + \sigma(\frac{Q_c^i(K_b^i)^T}{\sqrt{d_k}})V_b^i,
		\vspace{-1mm}
	\end{split}
\end{equation}
where $d_k$ is the channel length of projected keys for both $K_t^i$ and $K_b^i$. Similarly, We could calculate the aggregated feature $F_{b\_agg}^i$ for the other stream. Thus, final outputs of the $i^{th}$ CMB layer are formulated as:
\begin{equation}
	\small
	\begin{split}
		F_{t}^{i+1} = \mathcal{FFN}(\mathcal{LN}(F_{t}^{i} + F_{t\_agg}^i)) + (F_{t}^{i} + F_{t\_agg}^i),\\
		F_{b}^{i+1} = \mathcal{FFN}(\mathcal{LN}(F_{b}^{i} + F_{b\_agg}^i)) + (F_{b}^{i} + F_{b\_agg}^i),
		\vspace{-1mm}
	\end{split}
\end{equation}
where $\mathcal{LN}$ is layer normalization and $\mathcal{FFN}$ denotes multilayer perceptron. In experiments, we implement the hallucination module with $6$ RCMBs with depths $[6,8,8,8,8,6]$ and number of heads $[6,6,6,6,6,6]$, using an initial channel dimension of 96.

\begin{figure*}[!tb]
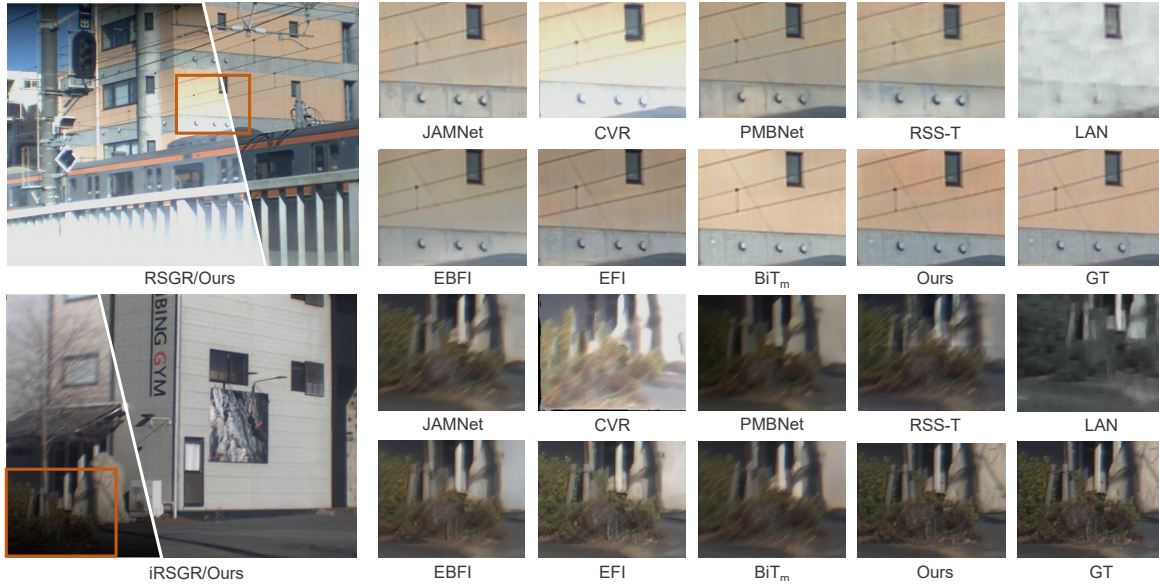

	\centering
	\mfigure{0.9}{exp/sota_comparison.pdf}
	\vspace{-2mm}
	\caption{
		Qualitative comparisons on RSGR-HDR. Our model outperforms the existing approaches with different settings in details recovery and dynamic range expansion.
	}
	\label{fig:sota_comparison}
	\vspace{-2mm}
\end{figure*}

\begin{table*}[!thp]
	\centering
	\setlength{\tabcolsep}{3pt}
	\caption{Quantitative comparisons on RSGR-HDR. We present experimental results on recovering a single latent HDR frame (HDR reconstruction) and an HDR sequence of length $3$, $5$, and $9$ (HDR photosequencing), both extracted within the exposure duration of the input frame. Performance is evaluated using mean PSNR/SSIM/LPIPS/TMQI. Dual reversed RSGR inputs are denoted as `RG-iRG'. `n$\cdot$RG' is the setting using $n$ neighboring RSGR frames and `RG-E' combines RSGR with event streams. Note that, LAN takes 5 consecutive LDR frames with alternate long and short exposures (denoted as '5$\cdot$LS').
	We also compute initial performance of inputs as RSGR and iRSGR.}
	\vspace{-2mm}
	\label{tab:compare_with_sota}
	\resizebox{\linewidth}{!}
	{
		\begin{tabular}{rcccccccc}
			\toprule
			\multirow{2}{*}{Method}& \multirow{2}{*}{Input} & \multirow{2}{*}{HDR Reconstruction} &\multicolumn{3}{c}{HDR Photosequencing} &\multirow{2}{*}{\begin{tabular}[c]{@{}l@{}}Time  \\ \quad(s)\end{tabular}}  & \multirow{2}{*}{\begin{tabular}[c]{@{}l@{}}Params\\    \quad(M)\end{tabular}} &  \multirow{2}{*}{\begin{tabular}[c]{@{}l@{}}FLOPs\\ \quad(G)\end{tabular}}  \\ \cmidrule(lr){4-6}
			
			& & &   $\times3$  &$\times5$     & $\times9$ &  &  &  \\ \midrule
			
			RSGR & \multirow{2}{*}{--} & 12.30/0.5450/0.2774/0.7157 & -- / -- / --  & -- / -- / -- & -- / -- / -- & -- & -- & --  \\ 
			
			iRSGR &   & 14.94/0.5961/0.2572/0.6993 & -- / -- / -- & -- / -- / -- & -- / -- / --  & -- & -- & -- \\ \midrule
			
			DSUN ~\cite{liu2020deep} & \multirow{8}{*}{\emph{2}$\cdot$\emph{RG}} & 19.44/0.7278/0.1663/0.7860 & -- / -- / --  & -- / -- / -- & -- / -- / -- & 0.847 & 3.900 & 225.0  \\ 
			
			JAMNet ~\cite{fan2023joint} &   & 21.39/0.7796/0.1397/0.8310 & -- / -- / -- & -- / -- / -- & -- / -- / --  & 0.690 & 4.730 & 51.13 \\
			
			RSSR ~\cite{fan2021inverting} & &  12.82/0.6099/0.2427/0.7924 &  12.75/0.5975/0.2566/0.7871  & 12.76/0.5970/0.2592/0.7895 & 12.72/0.5929/0.2639/0.7906 & 1.469  &26.03 & 42.67\\
			
			CVR ~\cite{fan2022context} &  & 13.04/0.5623/0.2898/0.7769  &  13.03/0.5609/0.2946/0.7565  & 13.04/0.5643/0.2963/0.7573  & 13.03/0.5621/0.3000/0.7578 & 1.497 & 42.70 & 101.1\\
			
			RIFE~\cite{huang2022real} &  & 19.80/0.7542/0.1911/0.8227  &  19.64/0.7486/0.1970/0.7998  & 19.67/0.7492/0.1981/0.8005  & 19.68/0.7496/0.1984/0.8008 & 0.358 & 54.82 & 71.15\\
			
			AfB~\cite{zhong2022animation} &  & 17.56/0.7285/0.2969/0.8342  &  17.41/0.7217/0.3060/0.8104  & 17.43/0.7228/0.3056/0.8112  & 17.44/0.7232/0.3054/0.8115 & 0.420 & 129.2 & 793.7 \\
			
			IFED~\cite{zhong2022bringing} &  & 12.30/0.4635/0.2723/0.7377  &  12.24/0.4579/0.2847/0.6831  & 12.25/0.4597/0.2835/0.6867  & 12.25/0.4607/0.2828/0.6883 & 0.394 & 10.81 & \textbf{29.58} \\
			
			PMBNet~\cite{ji2023rethinking}&  & 20.05/0.7593/0.1819/0.7961  &  19.98/0.7560/0.1838/0.7970  & 19.97/0.7556/0.1845/0.7980  & 19.97/0.7554/0.1850/0.7984 & 1.497 & 30.10 & 121.2\\  \midrule
			
			BiT~\cite{zhong2023blur} & \emph{3}$\cdot$\emph{RG} &  18.60/0.7437/0.1783/0.8111 &  18.59/0.7447/0.1884/0.7938  & 18.60/0.7452/0.1852/0.7944  & 18.61/0.7456/0.1833/0.7946 & 0.167 & 11.33 & 57.48 \\ \midrule
			
			ExpandNet~\cite{marnerides2018expandnet} & \multirow{2}{*}{\emph{1}$\cdot$\emph{RG}} & 16.26/0.6405/0.2956/0.7279 & -- / -- / --  & -- / -- / -- & -- / -- / --   & 0.587 & \textbf{0.457} & 63.17 \\
			
			RSS-T~\cite{ji2023single} &   & 19.46/0.7076/0.2024/0.7903 &  -- / -- / --   & -- / -- / -- & -- / -- / --  & 0.136 & 11.34 & 176.6 \\ \midrule
			
			DeMFI~\cite{oh2022demfi} & \emph{4}$\cdot$\emph{RG} &  14.54/0.3247/0.6156/0.6194 &  14.79/0.3894/0.5207/0.6404  & 14.73/0.3835/0.5282/0.6390  & 14.72/0.3763/0.5355/0.6477  & 0.388 &7.410 & 420.0\\ \midrule

			NGS~\cite{wang2022neural}&\emph{5}$\cdot$\emph{RG} & 19.17/0.6662/0.1804/0.7405 & -- / -- / --  & -- / -- / -- & -- / -- / --    & 1.555 & 8.670  & 1266\\ \midrule
			
			LAN~\cite{chung2023lan} &  \multirow{2}{*}{\emph{5}$\cdot$\emph{LS}} & 23.66/0.7122/0.3552/0.6582 &  -- / -- / --   & -- / -- / -- & -- / -- / --  & 0.243 & 7.320 & 949.1 \\ 
			
			HDRFlow~\cite{xu2024hdrflow} &   & 22.07/0.4464/0.3425/0.6712 & -- / -- / -- & -- / -- / -- & -- / -- / --  & \textbf{0.124} & 3.270 & 83.37 \\ \midrule
			
			EBFI~\cite{weng2023event}& \multirow{4}{*}{\emph{RG}-\emph{E}} & 18.25/0.7518/0.1464/0.8539 & 18.21/0.7490/0.1481/\textbf{0.8480}  & 18.21/0.7498/0.1474/0.8485 & 18.21/0.7503 /0.1469/0.8488  & 5.253 &5.680  & 432.2  \\
			
			EvUnroll~\cite{zhou2022evunroll}&   & 14.39/0.3929/0.3285/0.4815 &  14.38/0.3903/0.3299/0.4837   & 14.39/0.3909/0.3296/0.4841 & 14.39/0.3911/0.3294/0.4845  &  2.594 & 27.47 & 890.0\\
			
			EFI~\cite{lin2023event}&   & 18.61/0.8023/0.1664/0.8408 & 18.56/0.8006/0.1675/0.8443   & 18.57/0.8011/0.1671/0.8446 & 18.57/0.8014/0.1668/0.8447  &5.534  &16.60  & 977.2 \\ 
			
			UniINR~\cite{lu2024uniinr}&   & 20.75/0.8008/0.2173/0.8337 & 20.58/0.7880/0.2350/0.8299   & 20.60/0.7890/0.2335/0.8304 & 20.61/0.7898/0.2322/0.8307  &1.487  & 1.824  & 1366 \\ \midrule
			
			RIFE$_m$~\cite{huang2022real} & \multirow{4}{*}{\emph{RG}-\emph{iRG}} & 22.37/0.8445/0.1262/0.8486 & 22.12/0.8341 /0.1322/0.8441  & 22.15/0.8361/0.1322/0.8449 & 22.17/0.8371/0.1317/0.8453   & 0.358 & 54.82 & 71.15  \\
			
			IFED$_m$~\cite{zhong2022bringing}&  & 21.79/0.8166/0.1072/0.8509  &  21.49/0.8003/0.1172/0.8398  & 21.54/0.8033/0.1146/0.8415  & 21.57/0.8050/0.1131/0.8423  & 0.394 & 10.81 & \textbf{29.58} \\
			
			BiT$_m$~\cite{zhong2023blur}&  & 22.51/0.8341/0.1279/0.8512  & 22.32/0.8289/0.1319/0.8426  & 22.38/0.8303/0.1305/0.8434  & 22.40/0.8311/0.1298/0.8443  &0.167 & 11.33 & 57.48 \\
			
			Ours &  & \textbf{24.24/0.8679/0.0994/0.8543} &  \textbf{23.84/0.8562/0.1071/0.8480}   &  \textbf{23.91/0.8584/0.1058/0.8487}  & \textbf{23.95/0.8595/0.1053/0.8490} & 0.598 & 35.05 &183.1 \\   \bottomrule
		\end{tabular}
	}
	\vspace{-6mm}
\end{table*}

\paragraph{Tri-pivot Temporal Loss}
To generate perceptually natural video, avoiding the flickering is very essential. Since the correlation of reconstructed HDR frames is weakly constrained, we present tri-pivot temporal loss $\mathcal{L}_{temp}$ to improve structural coherence and temporal consistency. In specific, we sample three frame pivots that are temporally located at $i=\{0, \frac{N-1}{2},N-1\}$ from the reconstructed HDR video and enforce the difference maps among them to be similar to these of corresponding GTs using the Charbonnier function. On the other hand, we observe that discontinuity or inconsistency is normally caused by wrongly estimated edges and geometric structures. Therefore, instead of calculating temporal loss in sRGB space without distinguishing colors and contents, we conduct it on Y channel of YCrCb space in order to pay more attention to structures and details weakening the effects of color information.
The temporal loss for this enforcement is defined as:
\begin{equation}
	\small
	\begin{aligned}
		\Delta_{a,b}I &\triangleq [I^a]_Y - [I^b]_Y, \\
		\mathcal{L}_{temp}
		&= \sum_{(a,b)\in\mathcal{P}}
		\sqrt{
			\left\|
			\Delta_{a,b}H-\Delta_{a,b}G
			\right\|^2+\epsilon^2
		}.
	\end{aligned}
\end{equation}
where $\mathcal{P}=\{(N-1,0),(N-1,\frac{N-1}{2}),(\frac{N-1}{2},0)\}$. $[\cdot]_Y$ denotes extraction of the Y channel of YCrCb color space and $\epsilon$ is set as $10^{-3}$. $G$ represents groundtruth frame. While the reconstruction loss, $\mathcal{L}_{rec}$ is defined as:
\begin{equation}
	\small
	\begin{split}
		\mathcal{L}_{rec} = \sum_{i=0}^{N-1}\sqrt{\Vert H^i - G^i    \Vert^2 + \epsilon^2}.
	\end{split}
\end{equation}
In total, our training loss is $\mathcal{L} = \mathcal{L}_{rec} + \lambda \mathcal{L}_{temp}$, where $\lambda$ is set as 0.1 based on experiments to balance the reconstruction and temporal losses.

\section{Experiments}
\label{sec:experiments}



\begin{table}[!tp]
	\centering
	\setlength\tabcolsep{4pt}
	\caption{Model architecture ablation. RFA$_s$ and RFA$_m$ denote row-wise alignment with single scale and row-adaptive alignment with multiple scale. STB is Swin Transformer block. R, T$_R$ and T$_Y$ are reconstruction loss, temporal loss calculated in sRGB and YCrCb space. T$_W$ is warping-based temporal constraint.}
	\vspace{-2mm}
	\resizebox{\linewidth}{!}{
		\begin{tabular}{lcccccccccc} 
			\toprule
			& \multicolumn{3}{c}{Alignment} & \multicolumn{2}{c}{Transformer} & \multicolumn{4}{c}{Loss} & \multirow{2}{*}{PSNR / SSIM / LPIPS} \\
			\cmidrule(lr){2-4}
			\cmidrule(lr){5-6}
			\cmidrule(lr){7-10}
			& \small{Dense} & \small{RFA$_s$} & \small{RFA$_m$} & \small{STB} & \small{CMB}  &  \small{R} &  \small{R+T$_R$} & \small{R+T$_W$} & \small{R+T$_Y$} &  \\
			\midrule
			{$V_1$} &          &           &          &          &\checkmark&          &          &          &\checkmark&22.29 / 0.8428 / 0.1254 \\
			{$V_2$} &\checkmark&           &          &          &\checkmark&          &          &          &\checkmark&22.05 / 0.8299 / 0.1234 \\
			{$V_3$} &          &\checkmark &          &          &\checkmark&          &          &          &\checkmark&22.79 / 0.8383 / 0.1214 \\
			{$V_4$} &          &           &\checkmark&\checkmark&          &          &          &          &\checkmark&22.48 / 0.8372 / 0.1208 \\
			{$V_5$} &          &           &\checkmark&          &\checkmark&\checkmark&          &          &          &23.37 / 0.8500 / 0.1218 \\
			{$V_6$} &          &           &\checkmark&          &\checkmark&          &\checkmark&          &          &23.52 / 0.8525 / 0.1204 \\
			{$V_7$} &          &           &\checkmark&          &\checkmark&          &  &    \checkmark      &          &23.59 / 0.8426 /
			0.1483  \\
			\midrule
			\small{Ours} &          &           &\checkmark&          &\checkmark&          &          &  &\checkmark& \textbf{23.95 / 0.8595 / 0.1053 }\\
			\bottomrule
		\end{tabular}
	}
	\label{tab:ablation}
	\vspace{-3mm}
\end{table}

\begin{table}[!tp]
	\centering
	\setlength\tabcolsep{7pt}
	\caption{Row-shaped attention ablation. We conduct the comparisons to decide the optimal row shape for feature alignment.}
	\vspace{-2mm}
	\resizebox{\columnwidth}{!}{
		\begin{tabular}{llcc} 
			\toprule
			Row Size & Stride & Attention Type & {PSNR / SSIM / LPIPS} \\
			\midrule
			$r = [2]$ & $s = [1]$  &  \multirow{3}{*}{\begin{tabular}[c]{@{}c@{}} Row-wise\\ (Single Scale)\end{tabular}}     &22.10 / 0.8316 / 0.1353 \\
			$r = [3]$ & $s = [2]$  &         &22.61 / 0.8385 / 0.1276 \\
			$r = [5]$ & $s = [4]$  &         &22.79 / 0.8383 / 0.1214 \\
			\midrule
			$r = [2,3]$ &  $s = [1,2]$   &  \multirow{5}{*}{\begin{tabular}[c]{@{}c@{}}Row-adaptive \\(Multi Scale)\end{tabular}}         &23.68 / 0.8542 / 0.1159 \\
			$r = [3,5]$ &  $s = [3,5]$   &           &23.22 / 0.8567 / 0.1097 \\
			$r = [2,5]$ &  $s = [1,4]$   &           &\textbf{23.95 / 0.8595 / 0.1053} \\
			$r = [2,3,5]$ &  $s = [2,3,5]$   &           &23.65 / 0.8576 / 0.1056 \\
			$r = [2,3,5]$ &  $s = [1,2,4]$   &           &23.90 / 0.8556 / 0.1088 \\
			\bottomrule
		\end{tabular}
	}
	\label{tab:rowshape_attention}
	\vspace{-3mm}
\end{table}

\subsection{Comparison with SOTA methods}
\label{sec:comp_sota}
We aim to reconstruct a sequence of HDR frames from dual reversed RSGR views with intermediate degradation between RS effects and blur. Therefore, we compare our model with related SOTA methods from: 
\begin{itemize}[leftmargin=*]
	\vspace{-3mm}
	\item Blur interpolation and decomposition: BiT~\cite{zhong2023blur}, AfB~\cite{zhong2022animation} and DeMFI~\cite{oh2022demfi} approximate motions based on temporally adjacent blurry frames. We also include the typical interpolation method, RIFE~\cite{huang2022real} and more competitive setting by using the combination of blur frame and corresponding event streams, EBFI~\cite{weng2023event} and UniINR~\cite{lu2024uniinr}. 
	\vspace{-3mm}
	\item RS correction and interpolation: CVR~\cite{fan2022context}, RSSR~\cite{fan2021inverting}, DSUN~\cite{liu2020deep} and JAMNet~\cite{fan2023joint} take as input multiple RS frames while IFED~\cite{zhong2022bringing} depends on dual reversed RS views and EvUnroll~\cite{zhou2022evunroll} requires extra event information.
	\vspace{-3mm}
	\item RSGR undistortion and interpolation: NGS~\cite{wang2022neural} and RSS-T~\cite{ji2023single} reconstruct single sharp output while PMBNet~\cite{ji2023rethinking} can interpolate at arbitrary time. 
	\vspace{-3mm} 
	\item HDR reconstruction: LAN~\cite{chung2023lan}, HDRFlow~\cite{xu2024hdrflow} and ExpandNet~\cite{marnerides2018expandnet} follow convention by alternating exposures of input LDR frames. EFI~\cite{lin2023event} absorbs complementary information from the RS image and events to retrieve a HDR sequence.
	\vspace{-3mm}
\end{itemize}
To better demonstrate the superiority of our setup, we also adapt RIFE, IFED and BiT to dual RSGR inputs, which are denoted as RIFE$_m$, IFED$_m$, and BiT$_m,$ respectively. As a contrast, RIFE, IFED and BiT represent models using consecutive RSGR frames. 
Importantly, following prior work~\cite{mertens2007exposure,lin2023event,chen2025ultrafusion}, we adopt an exposure-fusion strategy for HDR composition. Unlike conventional pipelines that reconstruct a linear HDR image followed by tone mapping, we directly output tone-mapped LDR results, avoiding error accumulation~\cite{chen2025ultrafusion}. Accordingly, methods are evaluated using PSNR, SSIM, LPIPS and TMQI~\cite{yeganeh2012objective}, rather than linear-HDR metrics such as PSNR-PU, SSIM-$\mu$ or HDR-VDP.

After retraining all models on RSGR-HDR, we present experimental results in~\tabref{compare_with_sota}. Our method outperforms all existing approaches across diverse setups by a large margin (at least 1.5 dB on PSNR). Overall, algorithms designed for motion degradation fail under extreme overexposure or underexposure conditions, while HDR methods cannot handle motion distortions in dynamic scenes. The significant improvements of RIFE$_m$, IFED$_m$, and BiT$_m$ over their corresponding baselines (RIFE, IFED, and BiT) clearly demonstrate the effectiveness and generalizability of our dual reversed setup. The qualitative comparison in~\figref{sota_comparison} further validates that our method can simultaneously expand dynamic range and reconstruct visual details from motion degradation. We further present the inference-time computational cost of all models in~\tabref{compare_with_sota}, measured on a single NVIDIA GeForce RTX 4090 with an input size of $640 \times 480$. The results demonstrate that our model attains the highest perceptual and fidelity scores while remaining competitively efficient.

To better substantiate the ability of our model in temporal consistency and local details recovery, we also provide consecutive recovered frames and video demos in~\secref{video_recon} and supplemental material, respectively.

\begin{figure}[!tb]
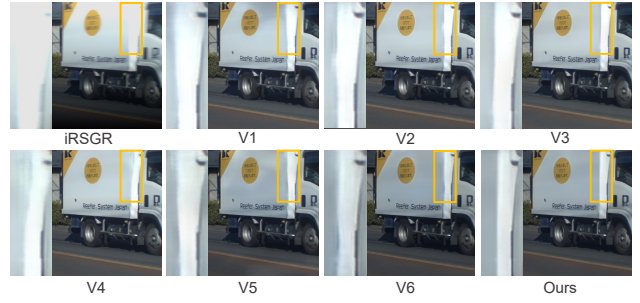

	\centering
	\mfigure{1}{exp/ablation_study_v3.pdf}
	\vspace{-6mm}
	\caption{
		Visual results of model architecture ablation. We can observe that each key component contributes to the final results.
	}
	\label{fig:ablation}
	\vspace{-6mm}
\end{figure}

\subsection{Ablation Study}
To assess the efficacy of proposed structure in our model, we conduct ablation studies as depicted in~\tabref{ablation}. We first remove feature alignment ($V_1$), and then successively replace our row-adaptive strategy with the dense and row-wise counterparts ($V_2$ and $V_3$). The performance significantly drops in all metrics. We also observe that the Swin Transformer block ($V_4$) fails to capture the complementarity between RSGR-iRSGR views. Although incorporating temporal loss in the sRGB space leads to certain improvements ($V_5$ vs.~$V_6$), the gains remain limited. In contrast, our proposed tri-pivot loss contributes substantially to the final performance. We also compared with frequently used warping-based temporal constraint~\cite{lai2018learning}.~\figref{ablation} demonstrates consistent findings from qualitative aspect. Moreover, we perform ablation study on row shapes used in feature alignment to explore the optimal hyper-parameters. As shown in~\tabref{rowshape_attention}, our model is retrained with different combinations of row size $r$ and stride $s$, which further proves the effectiveness of our tailored design.

\begin{table}[!t]
	\centering
	\caption{Quantitative comparisons on misaligned RSGR-iRSGR view. `Shift-$n$' denotes misalignment with maximal offsets $n$.}
	\vspace{-1mm}
	\label{tab:visual_shift}
	\resizebox{\linewidth}{!}
	{
		\begin{tabular}{rccccccccc}
			\toprule
			\multirow{2}{*}{Method} & \multicolumn{3}{c}{$\times$3} & \multicolumn{3}{c}{$\times$5} &  \multicolumn{3}{c}{$\times$9} \\ \cmidrule(lr){2-4}\cmidrule(lr){5-7}\cmidrule(lr){8-10}
			
			& PSNR     & SSIM     & LPIPS    & PSNR    & SSIM    & LPIPS & PSNR    & SSIM    & LPIPS   \\ \midrule
			
			Shift-$4$ & \textbf{23.88}  & 0.8483 & 0.1273  &  \textbf{23.94} &  0.8501 & 0.1262 & \textbf{23.97} & 0.8511 & 0.1255 \\
			
			Shift-$6$ & 23.81 & 0.8536 & 0.1202 & 23.88 & 0.8555 & 0.1190 & 23.91 & 0.8563 & 0.1185\\		 	
			
			Shift-$8$ & 23.79 & 0.8464 & 0.1275 & 23.86 & 0.8483 & 0.1262 & 23.89 & 0.8493 & 0.1256 \\
			
			Ours &   23.84 & \textbf{0.8562} & \textbf{0.1071}  & 23.91  & \textbf{0.8584}  &  \textbf{0.1058} & 23.95 & \textbf{0.8595} & \textbf{0.1053} \\ \bottomrule
		\end{tabular}
	}
	\vspace{-2mm}
\end{table}

\begin{figure}[!t]
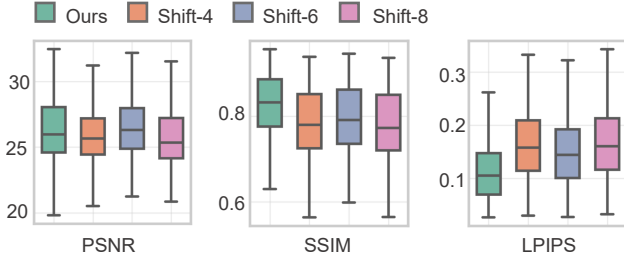

	\mfigure{1}{exp/misaligned_drawing_v3.pdf}
	\vspace{-6mm}
	\caption{
		Performance distribution of our method and misaligned variants(`Shift-$4$', `Shift-$6$' and `Shift-$8$') under a selected scene. The box-plot further intuitively validate robustness of our solution to exposure convergence and visual shifts under fast motion.
	}
	\label{fig:visual_shift}
	\vspace{-1mm}
\end{figure}

\begin{table}[!t]
	\centering
	\setlength{\tabcolsep}{2pt}
	\caption{Quantitative comparisons on GOPRO-Dual. We follow the same evaluation protocol used in our experiments on RSGR-HDR. Performance is evaluated using mean PSNR/SSIM/LPIPS/TMQI.
		} 
	\vspace{-1mm}
	\label{tab:synthetic_comparison}
	\resizebox{1\linewidth}{!}
	{
		\begin{tabular}{rccc}
			\toprule
			{Method}& {Input} & {HDR Reconstruction} &{HDR Photosequencing} ($\times9$) \\ \midrule
			
			DSUN ~\cite{liu2020deep}& \multirow{8}{*}{\emph{2}$\cdot$\emph{RG}} & 19.92 / 0.7850 / 0.1347 / 0.8691& -- / -- / --   \\ 
			
			JAMNet ~\cite{fan2023joint}& & 25.16 / 0.8797 / 0.0940 / 0.9186 & -- / -- / --   \\ 
			
			RSSR ~\cite{fan2021inverting}&  & 12.21 / 0.6535 / 0.2394 / 0.8164 &  13.02 / 0.6908 / 0.1956 / 0.8060  \\ 
			
			CVR ~\cite{fan2022context}&   & 13.29 / 0.6886 / 0.2291 / 0.8211 &   12.98 / 0.6699 / 0.2062 / 0.7978  \\
			
			RIFE~\cite{huang2022real} & &  20.97 / 0.8327 / 0.1646 / 0.9412 &   20.96 / 0.8335 / 0.1710 / 0.8910 \\
			
			AfB~\cite{zhong2022animation} & &  20.64 / 0.7021 / 0.3080 / 0.9354 &   20.84 / 0.7204 / 0.3092 / 0.8553 \\
			IFED~\cite{zhong2022bringing}&  & 23.96 / 0.8447 / 0.0949 / 0.9431 &  24.24 / 0.8434 / 0.1000 / 0.8989 \\
			
			PMBNet~\cite{ji2023rethinking}&  & 25.42 / 0.9009 / 0.1061 / 0.9252 &   25.11 / 0.8923 / 0.1160 / 0.9180 \\ \midrule
			
			BiT~\cite{zhong2023blur} & \emph{3}$\cdot$\emph{RG} &  23.96 / 0.8673 / 0.1034 / 0.9392 & 24.35 / 0.8921 / 0.0882 / 0.9273 \\ \midrule
			
			ExpandNet~\cite{marnerides2018expandnet} & \multirow{2}{*}{\emph{1}$\cdot$\emph{RG}} & 17.59 / 0.7410 / 0.2702 / 0.8274 & -- / -- / --    \\
			
			RSS-T~\cite{ji2023single} &   & 23.00 / 0.9229 / 0.0756 / 0.9316 & -- / -- / --  \\ \midrule
			
			DeMFI~\cite{oh2022demfi} & \emph{4}$\cdot$\emph{RG} &  21.06 / 0.8520 / 0.1324 / 0.9203 &   20.59 / 0.8191 / 0.2014 / 0.8910 \\ \midrule

			NGS~\cite{wang2022neural}& \emph{5}$\cdot$\emph{RG} & 31.04 / 0.9645 / 0.0343 / 0.9457 & -- / -- / --    \\  \midrule
			
			LAN~\cite{chung2023lan} &  \multirow{2}{*}{\emph{5}$\cdot$\emph{LS}} & 34.38 / 0.9416 / 0.0589 / 0.7003 &   -- / -- / --  \\ 
			HDRFlow~\cite{xu2024hdrflow} & & 21.17 / 0.4041 / 0.4663 / 0.6520 & -- / -- / --   \\ \midrule
			
			EBFI~\cite{weng2023event}& \multirow{4}{*}{\emph{RG}-\emph{E}} & 20.82 / 0.8864 / 0.1034 / 0.9354 & 20.77 / 0.8844 / 0.1034 / 0.9233   \\
			
			EvUnroll~\cite{zhou2022evunroll}&   & 29.27 / 0.9372 / 0.0496 / 0.9446 &  29.17 / 0.9345 / 0.0509 / 0.9351 \\
			
			EFI~\cite{lin2023event}&   & 20.45 / 0.8905 / 0.1005 / 0.9427 & 20.41 / 0.8874 / 0.1027 / 0.9373 \\
			UniINR~\cite{lu2024uniinr} &   & 24.21 / 0.9124 / 0.1090 / 0.9335 & 23.31 / 0.8448 / 0.1794 / 0.9111 \\ \midrule
			
			RIFE$_m$~\cite{huang2022real} & \multirow{3}{*}{\emph{RG}-\emph{iRG}} & 32.39 / 0.9591 / 0.0659 / 0.9544 &  32.49 / 0.9516 / 0.0712 / 0.9412   \\
			
			IFED$_m$~\cite{zhong2022bringing}&  & 32.34 / 0.9625 / \textbf{0.0291} / 0.9542  &   32.05 / 0.9509 / \textbf{0.0327} / 0.9427 \\
			
			BiT$_m$~\cite{zhong2023blur}&  & 34.17 / 0.9733 / 0.0400 / 0.9538 &  33.53 / 0.9637 / 0.0449 / 0.9444 \\
			
			Ours &  & \textbf{35.88} / \textbf{0.9795} / 0.0293 / \textbf{0.9546} & \textbf{35.55} / \textbf{0.9738} / 0.0364 / \textbf{0.9474}  \\ \bottomrule
		\end{tabular}
	}
	\vspace{-4mm}
\end{table}

\subsection{Visual Shifts and Convergent Exposures}
 \label{sec:visual_shift}
 
 As discussed in~\secref{intro}, the exposure convergence in the middle rows of RSGR-iRSGR views may complicate HDR expansion, but missing contrast can be compensated by our model. On the other hand, random visual shifts between corresponding rows are inevitable due to physical misalignments and fast motion. Such shifts, in turn, effectively mitigate the exposure-convergence issue. Meanwhile, our dual setting normally equips some tolerance to these constrained visual shifts~\cite{zhong2022bringing,ji2024motion}. To prove this, we randomly shift the RSGR view in the image space along the horizontal and vertical axes, with maximal translational offsets as $4, 6, 8$ pixels, respectively. We then retrain our model using these misaligned pairs and present comparisons in \tabref{visual_shift} and \figref{visual_shift}. The results reveal no significant performance drop, validating that our dual-view setup and customized model remain robust to dynamic scenes, where motion-induced spatial displacement helps alleviate central exposure convergence and limits its impact on reconstruction quality.

\begin{figure}[!tb]
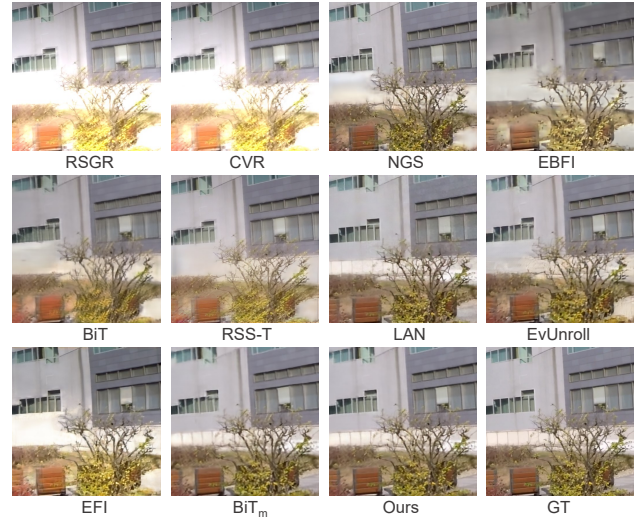

	\centering
	\mfigure{1}{exp/sota_comparison_goprovfi_v1.pdf}
	\vspace{-6mm}
	\caption{
		Qualitative comparisons on GOPRO-Dual. Our model outperforms existing approaches with different settings in details recovery and dynamic range expansion. Best viewed with zoom.
	}
	\label{fig:sota_comparison_goprovfi}
	\vspace{-6mm}
\end{figure}

\subsection{Comparison on Synthetic Data}
\label{sec:experiments_synthesic}
We also perform experiments on the synthetic dataset GOPRO-Dual to further demonstrate the conclusions drawn from real-data RSGR-HDR. Quantitative and qualitative results are presented in~\tabref{synthetic_comparison} and ~\figref{sota_comparison_goprovfi}. The dataset synthesizing process and more experimental results are presented in~\secref{data_training} and ~\secref{more_results}.

\section{Conclusion}
\label{sec:conclusion}

In this paper, we present a new task of HDR photosequencing from dual reversed RSGR inputs, addressing the inherent limitations of RSGR in frame rate and dynamic range. By exploiting the row-wise complementary property, our framework enables robust feature alignment and dynamic range expansion even under large motions. We also construct an optical system to collect dual reversed RSGR observations with sharp HDR sequences, and develop a network with row-adaptive alignment, mix-attention hallucination, and tri-pivot structural supervision, achieving improved detail recovery and reduced temporal flickering. Extensive experiments on both synthetic and real data validate the effectiveness of our proposed setting and model, demonstrating clear advantages over existing solutions.

\bibliography{example_paper}

@article{kronander2014unified,
	title={A unified framework for multi-sensor HDR video reconstruction},
	author={Kronander, Joel and Gustavson, Stefan and Bonnet, Gerhard and Ynnerman, Anders and Unger, Jonas},
	journal={Signal Processing: Image Communication},
	volume={29},
	number={2},
	pages={203--215},
	year={2014},
	publisher={Elsevier}
}

@inproceedings{su2017deep,
	title     = {Deep Video Deblurring for Hand-Held Cameras},
	author    = {Su, Shuochen and Delbracio, Mauricio and Wang, Jue and Sapiro, Guillermo and Heidrich, Wolfgang and Wang, Oliver},
	booktitle = {Proceedings of the IEEE Conference on Computer Vision and Pattern Recognition},
	pages     = {237--246},
	year      = {2017}
}

@inproceedings{zhao2025tree,
	title={Tree-NeRV: Efficient Non-Uniform Sampling for Neural Video Representation via Tree-Structured Feature Grids},
	author={Zhao, Jiancheng and Zhan, Yifan and Zhu, Qingtian and Ma, Mingze and Niu, Muyao and Wan, Zunian and Ji, Xiang and Zheng, Yinqiang},
	booktitle={Proceedings of the IEEE/CVF International Conference on Computer Vision},
	pages={15076--15085},
	year={2025}
}

@inproceedings{niu2024rs,
	title={Rs-nerf: Neural radiance fields from rolling shutter images},
	author={Niu, Muyao and Chen, Tong and Zhan, Yifan and Li, Zhuoxiao and Ji, Xiang and Zheng, Yinqiang},
	booktitle={European Conference on Computer Vision},
	pages={163--180},
	year={2024},
	organization={Springer}
}

@article{ji2026moment,
	title={Moment-Reenacting: Inverse Motion Degradation With Cross-Shutter Guidance},
	author={Ji, Xiang and Lin, Guixu and Yin, Zhengwei and Zhao, Jiancheng and Zheng, Yinqiang},
	journal={IEEE Transactions on Pattern Analysis and Machine Intelligence},
	year={2026},
	publisher={IEEE}
}

@inproceedings{xu2024hdrflow,
	title={Hdrflow: Real-time hdr video reconstruction with large motions},
	author={Xu, Gangwei and Wang, Yujin and Gu, Jinwei and Xue, Tianfan and Yang, Xin},
	booktitle={Proceedings of the IEEE/CVF Conference on Computer Vision and Pattern Recognition},
	pages={24851--24860},
	year={2024}
}

@inproceedings{lu2024uniinr,
	title={Uniinr: Event-guided unified rolling shutter correction, deblurring, and interpolation},
	author={Lu, Yunfan and Liang, Guoqiang and Wang, Yusheng and Wang, Lin and Xiong, Hui},
	booktitle={European Conference on Computer Vision},
	pages={1--20},
	year={2024},
	organization={Springer}
}

@inproceedings{lai2018learning,
	title={Learning blind video temporal consistency},
	author={Lai, Wei-Sheng and Huang, Jia-Bin and Wang, Oliver and Shechtman, Eli and Yumer, Ersin and Yang, Ming-Hsuan},
	booktitle={Proceedings of the European conference on computer vision (ECCV)},
	pages={170--185},
	year={2018}
}

@inproceedings{rim2020real,
	title={Real-world blur dataset for learning and benchmarking deblurring algorithms},
	author={Rim, Jaesung and Lee, Haeyun and Won, Jucheol and Cho, Sunghyun},
	booktitle={European conference on computer vision},
	pages={184--201},
	year={2020},
	organization={Springer}
}

@inproceedings{mertens2007exposure,
	title={Exposure fusion},
	author={Mertens, Tom and Kautz, Jan and Van Reeth, Frank},
	booktitle={15th Pacific Conference on Computer Graphics and Applications (PG'07)},
	pages={382--390},
	year={2007},
	organization={IEEE}
}

@article{yeganeh2012objective,
	title={Objective quality assessment of tone-mapped images},
	author={Yeganeh, Hojatollah and Wang, Zhou},
	journal={IEEE Transactions on Image processing},
	volume={22},
	number={2},
	pages={657--667},
	year={2012},
	publisher={IEEE}
}

@inproceedings{chen2025ultrafusion,
	title={UltraFusion: Ultra high dynamic imaging using exposure fusion},
	author={Chen, Zixuan and Wang, Yujin and Cai, Xin and You, Zhiyuan and Lu, Zheming and Zhang, Fan and Guo, Shi and Xue, Tianfan},
	booktitle={Proceedings of the Computer Vision and Pattern Recognition Conference},
	pages={16111--16121},
	year={2025}
}

@inproceedings{shi2021stereo,
	title={Stereo waterdrop removal with row-wise dilated attention},
	author={Shi, Zifan and Fan, Na and Yeung, Dit-Yan and Chen, Qifeng},
	booktitle={2021 IEEE/RSJ International Conference on Intelligent Robots and Systems (IROS)},
	pages={3829--3836},
	year={2021},
	organization={IEEE}
}

@article{li2024era3d,
	title={Era3d: High-resolution multiview diffusion using efficient row-wise attention},
	author={Li, Peng and Liu, Yuan and Long, Xiaoxiao and Zhang, Feihu and Lin, Cheng and Li, Mengfei and Qi, Xingqun and Zhang, Shanghang and Xue, Wei and Luo, Wenhan and others},
	journal={Advances in Neural Information Processing Systems},
	volume={37},
	pages={55975--56000},
	year={2024}
}

@article{ye2020drm,
	title={DRM-SLAM: Towards dense reconstruction of monocular SLAM with scene depth fusion},
	author={Ye, Xinchen and Ji, Xiang and Sun, Baoli and Chen, Shenglun and Wang, Zhihui and Li, Haojie},
	journal={Neurocomputing},
	volume={396},
	pages={76--91},
	year={2020},
	publisher={Elsevier}
}

@inproceedings{ji2018dense,
	title={Dense reconstruction from monocular SLAM with fusion of sparse map-points and CNN-inferred depth},
	author={Ji, Xiang and Ye, Xinchen and Xu, Hongcan and Li, Haojie},
	booktitle={2018 IEEE International Conference on Multimedia and Expo (ICME)},
	pages={1--6},
	year={2018},
	organization={IEEE}
}

@inproceedings{schubert2019rolling,
	title={Rolling-shutter modelling for direct visual-inertial odometry},
	author={Schubert, David and Demmel, Nikolaus and von Stumberg, Lukas and Usenko, Vladyslav and Cremers, Daniel},
	booktitle={2019 IEEE/RSJ International Conference on Intelligent Robots and Systems (IROS)},
	pages={2462--2469},
	year={2019},
	organization={IEEE}
}

@inproceedings{albl2020two,
	title={From two rolling shutters to one global shutter},
	author={Albl, Cenek and Kukelova, Zuzana and Larsson, Viktor and Polic, Michal and Pajdla, Tomas and Schindler, Konrad},
	booktitle={Proceedings of the IEEE/CVF Conference on Computer Vision and Pattern Recognition},
	pages={2505--2513},
	year={2020}
}

@inproceedings{zhuang2017rolling,
	title={Rolling-shutter-aware differential sfm and image rectification},
	author={Zhuang, Bingbing and Cheong, Loong-Fah and Hee Lee, Gim},
	booktitle={Proceedings of the IEEE International Conference on Computer Vision},
	pages={948--956},
	year={2017}
}

@inproceedings{lao2018robust,
	title={A robust method for strong rolling shutter effects correction using lines with automatic feature selection},
	author={Lao, Yizhen and Ait-Aider, Omar},
	booktitle={Proceedings of the IEEE Conference on Computer Vision and Pattern Recognition},
	pages={4795--4803},
	year={2018}
}

@inproceedings{purkait2017rolling,
	title={Rolling shutter correction in manhattan world},
	author={Purkait, Pulak and Zach, Christopher and Leonardis, Ales},
	booktitle={Proceedings of the IEEE International Conference on Computer Vision},
	pages={882--890},
	year={2017}
}

@inproceedings{rengarajan2016bows,
	title={From bows to arrows: Rolling shutter rectification of urban scenes},
	author={Rengarajan, Vijay and Rajagopalan, Ambasamudram N and Aravind, Rangarajan},
	booktitle={Proceedings of the IEEE Conference on Computer Vision and Pattern Recognition},
	pages={2773--2781},
	year={2016}
}

@inproceedings{vasu2018occlusion,
	title={Occlusion-aware rolling shutter rectification of 3d scenes},
	author={Vasu, Subeesh and Rajagopalan, AN and others},
	booktitle={Proceedings of the IEEE Conference on Computer Vision and Pattern Recognition},
	pages={636--645},
	year={2018}
}

@inproceedings{zhuang2019learning,
	title={Learning structure-and-motion-aware rolling shutter correction},
	author={Zhuang, Bingbing and Tran, Quoc-Huy and Ji, Pan and Cheong, Loong-Fah and Chandraker, Manmohan},
	booktitle={Proceedings of the IEEE/CVF Conference on Computer Vision and Pattern Recognition},
	pages={4551--4560},
	year={2019}
}

@inproceedings{liu2020deep,
	title={Deep shutter unrolling network},
	author={Liu, Peidong and Cui, Zhaopeng and Larsson, Viktor and Pollefeys, Marc},
	booktitle={Proceedings of the IEEE/CVF Conference on Computer Vision and Pattern Recognition},
	pages={5941--5949},
	year={2020}
}

@article{ringaby2012efficient,
	title={Efficient video rectification and stabilisation for cell-phones},
	author={Ringaby, Erik and Forss{\'e}n, Per-Erik},
	journal={International journal of computer vision},
	volume={96},
	pages={335--352},
	year={2012},
	publisher={Springer}
}

@inproceedings{yan2023deep,
	title={Deep Homography Mixture for Single Image Rolling Shutter Correction},
	author={Yan, Weilong and Tan, Robby T and Zeng, Bing and Liu, Shuaicheng},
	booktitle={Proceedings of the IEEE/CVF International Conference on Computer Vision},
	pages={9868--9877},
	year={2023}
}

@inproceedings{fan2023joint,
	title={Joint Appearance and Motion Learning for Efficient Rolling Shutter Correction},
	author={Fan, Bin and Mao, Yuxin and Dai, Yuchao and Wan, Zhexiong and Liu, Qi},
	booktitle={Proceedings of the IEEE/CVF Conference on Computer Vision and Pattern Recognition},
	pages={5671--5681},
	year={2023}
}

@article{qu2023towards,
	title={Towards Nonlinear-Motion-Aware and Occlusion-Robust Rolling Shutter Correction},
	author={Qu, Delin and Lao, Yizhen and Wang, Zhigang and Wang, Dong and Zhao, Bin and Li, Xuelong},
	journal={arXiv preprint arXiv:2303.18125},
	year={2023}
}

@inproceedings{fan2021inverting,
	title={Inverting a rolling shutter camera: bring rolling shutter images to high framerate global shutter video},
	author={Fan, Bin and Dai, Yuchao},
	booktitle={Proceedings of the IEEE/CVF International Conference on Computer Vision},
	pages={4228--4237},
	year={2021}
}

@inproceedings{fan2022context,
	title={Context-aware video reconstruction for rolling shutter cameras},
	author={Fan, Bin and Dai, Yuchao and Zhang, Zhiyuan and Liu, Qi and He, Mingyi},
	booktitle={Proceedings of the IEEE/CVF Conference on Computer Vision and Pattern Recognition},
	pages={17572--17582},
	year={2022}
}

@inproceedings{zhong2022bringing,
	title={Bringing rolling shutter images alive with dual reversed distortion},
	author={Zhong, Zhihang and Cao, Mingdeng and Sun, Xiao and Wu, Zhirong and Zhou, Zhongyi and Zheng, Yinqiang and Lin, Stephen and Sato, Imari},
	booktitle={European Conference on Computer Vision},
	pages={233--249},
	year={2022},
	organization={Springer}
}

@inproceedings{wang2022neural,
	title={Neural global shutter: Learn to restore video from a rolling shutter camera with global reset feature},
	author={Wang, Zhixiang and Ji, Xiang and Huang, Jia-Bin and Satoh, Shin'ichi and Zhou, Xiao and Zheng, Yinqiang},
	booktitle={Proceedings of the IEEE/CVF Conference on Computer Vision and Pattern Recognition},
	pages={17794--17803},
	year={2022}
}

@inproceedings{ji2023single,
	title={Single Image Deblurring with Row-dependent Blur Magnitude},
	author={Ji, Xiang and Wang, Zhixiang and Satoh, Shin'ichi and Zheng, Yinqiang},
	booktitle={Proceedings of the IEEE/CVF International Conference on Computer Vision},
	pages={12269--12280},
	year={2023}
}

@inproceedings{chen2021hdr,
	title={HDR video reconstruction: A coarse-to-fine network and a real-world benchmark dataset},
	author={Chen, Guanying and Chen, Chaofeng and Guo, Shi and Liang, Zhetong and Wong, Kwan-Yee K and Zhang, Lei},
	booktitle={Proceedings of the IEEE/CVF international conference on computer vision},
	pages={2502--2511},
	year={2021}
}

@inproceedings{chung2023lan,
	title={LAN-HDR: Luminance-based Alignment Network for High Dynamic Range Video Reconstruction},
	author={Chung, Haesoo and Cho, Nam Ik},
	booktitle={Proceedings of the IEEE/CVF International Conference on Computer Vision},
	pages={12760--12769},
	year={2023}
}

@inproceedings{liu2022ghost,
	title={Ghost-free high dynamic range imaging with context-aware transformer},
	author={Liu, Zhen and Wang, Yinglong and Zeng, Bing and Liu, Shuaicheng},
	booktitle={European Conference on Computer Vision},
	pages={344--360},
	year={2022},
	organization={Springer}
}

@InProceedings{ji2024motion,
	author    = {Ji, Xiang and Jiang, Haiyang and Zheng, Yinqiang},
	title     = {Motion Blur Decomposition with Cross-shutter Guidance},
	booktitle = {Proceedings of the IEEE/CVF Conference on Computer Vision and Pattern Recognition (CVPR)},
	month     = {June},
	year      = {2024},
	pages     = {12534-12543}
}

@inproceedings{sun2018pwc,
	title={Pwc-net: Cnns for optical flow using pyramid, warping, and cost volume},
	author={Sun, Deqing and Yang, Xiaodong and Liu, Ming-Yu and Kautz, Jan},
	booktitle={Proceedings of the IEEE conference on computer vision and pattern recognition},
	pages={8934--8943},
	year={2018}
}

@inproceedings{khan2006ghost,
	title={Ghost removal in high dynamic range images},
	author={Khan, Erum Arif and Akyuz, Ahmet Oguz and Reinhard, Erik},
	booktitle={2006 International Conference on Image Processing},
	pages={2005--2008},
	year={2006},
	organization={IEEE}
}

@inproceedings{heo2010ghost,
	title={Ghost-free high dynamic range imaging},
	author={Heo, Yong Seok and Lee, Kyoung Mu and Lee, Sang Uk and Moon, Youngsu and Cha, Joonhyuk},
	booktitle={Asian Conference on Computer Vision},
	pages={486--500},
	year={2010},
	organization={Springer}
}

@article{zhang2011gradient,
	title={Gradient-directed multiexposure composition},
	author={Zhang, Wei and Cham, Wai-Kuen},
	journal={IEEE Transactions on Image Processing},
	volume={21},
	number={4},
	pages={2318--2323},
	year={2011},
	publisher={IEEE}
}

@article{oh2014robust,
	title={Robust high dynamic range imaging by rank minimization},
	author={Oh, Tae-Hyun and Lee, Joon-Young and Tai, Yu-Wing and Kweon, In So},
	journal={IEEE transactions on pattern analysis and machine intelligence},
	volume={37},
	number={6},
	pages={1219--1232},
	year={2014},
	publisher={IEEE}
}

@article{lee2014ghost,
	title={Ghost-free high dynamic range imaging via rank minimization},
	author={Lee, Chul and Li, Yuelong and Monga, Vishal},
	journal={IEEE signal processing letters},
	volume={21},
	number={9},
	pages={1045--1049},
	year={2014},
	publisher={IEEE}
}

@article{kang2003high,
	title={High dynamic range video},
	author={Kang, Sing Bing and Uyttendaele, Matthew and Winder, Simon and Szeliski, Richard},
	journal={ACM Transactions On Graphics (TOG)},
	volume={22},
	number={3},
	pages={319--325},
	year={2003},
	publisher={ACM New York, NY, USA}
}

@inproceedings{zimmer2011freehand,
	title={Freehand HDR imaging of moving scenes with simultaneous resolution enhancement},
	author={Zimmer, Henning and Bruhn, Andr{\'e}s and Weickert, Joachim},
	booktitle={Computer Graphics Forum},
	volume={30},
	number={2},
	pages={405--414},
	year={2011},
	organization={Wiley Online Library}
}

@article{sen2012robust,
	title={Robust patch-based hdr reconstruction of dynamic scenes.},
	author={Sen, Pradeep and Kalantari, Nima Khademi and Yaesoubi, Maziar and Darabi, Soheil and Goldman, Dan B and Shechtman, Eli},
	journal={ACM Trans. Graph.},
	volume={31},
	number={6},
	pages={203--1},
	year={2012}
}

@inproceedings{hu2013hdr,
	title={HDR deghosting: How to deal with saturation?},
	author={Hu, Jun and Gallo, Orazio and Pulli, Kari and Sun, Xiaobai},
	booktitle={Proceedings of the IEEE conference on computer vision and pattern recognition},
	pages={1163--1170},
	year={2013}
}

@inproceedings{chen2023learning,
	title={Learning continuous exposure value representations for single-image hdr reconstruction},
	author={Chen, Su-Kai and Yen, Hung-Lin and Liu, Yu-Lun and Chen, Min-Hung and Hu, Hou-Ning and Peng, Wen-Hsiao and Lin, Yen-Yu},
	booktitle={Proceedings of the IEEE/CVF International Conference on Computer Vision},
	pages={12990--13000},
	year={2023}
}

@inproceedings{marnerides2018expandnet,
	title={Expandnet: A deep convolutional neural network for high dynamic range expansion from low dynamic range content},
	author={Marnerides, Demetris and Bashford-Rogers, Thomas and Hatchett, Jonathan and Debattista, Kurt},
	booktitle={Computer Graphics Forum},
	volume={37},
	number={2},
	pages={37--49},
	year={2018},
	organization={Wiley Online Library}
}

@inproceedings{liu2020single,
	title={Single-image HDR reconstruction by learning to reverse the camera pipeline},
	author={Liu, Yu-Lun and Lai, Wei-Sheng and Chen, Yu-Sheng and Kao, Yi-Lung and Yang, Ming-Hsuan and Chuang, Yung-Yu and Huang, Jia-Bin},
	booktitle={Proceedings of the IEEE/CVF conference on computer vision and pattern recognition},
	pages={1651--1660},
	year={2020}
}

@article{kalantari2017deep,
	title={Deep high dynamic range imaging of dynamic scenes.},
	author={Kalantari, Nima Khademi and Ramamoorthi, Ravi and others},
	journal={ACM Trans. Graph.},
	volume={36},
	number={4},
	pages={144--1},
	year={2017}
}

@article{cai2018learning,
	title={Learning a deep single image contrast enhancer from multi-exposure images},
	author={Cai, Jianrui and Gu, Shuhang and Zhang, Lei},
	journal={IEEE Transactions on Image Processing},
	volume={27},
	number={4},
	pages={2049--2062},
	year={2018},
	publisher={IEEE}
}

@inproceedings{wu2018deep,
	title={Deep high dynamic range imaging with large foreground motions},
	author={Wu, Shangzhe and Xu, Jiarui and Tai, Yu-Wing and Tang, Chi-Keung},
	booktitle={Proceedings of the European Conference on Computer Vision (ECCV)},
	pages={117--132},
	year={2018}
}

@inproceedings{yan2019attention,
	title={Attention-guided network for ghost-free high dynamic range imaging},
	author={Yan, Qingsen and Gong, Dong and Shi, Qinfeng and Hengel, Anton van den and Shen, Chunhua and Reid, Ian and Zhang, Yanning},
	booktitle={Proceedings of the IEEE/CVF Conference on Computer Vision and Pattern Recognition},
	pages={1751--1760},
	year={2019}
}

@article{yan2020deep,
	title={Deep HDR imaging via a non-local network},
	author={Yan, Qingsen and Zhang, Lei and Liu, Yu and Zhu, Yu and Sun, Jinqiu and Shi, Qinfeng and Zhang, Yanning},
	journal={IEEE Transactions on Image Processing},
	volume={29},
	pages={4308--4322},
	year={2020},
	publisher={IEEE}
}

@article{niu2021hdr,
	title={HDR-GAN: HDR image reconstruction from multi-exposed LDR images with large motions},
	author={Niu, Yuzhen and Wu, Jianbin and Liu, Wenxi and Guo, Wenzhong and Lau, Rynson WH},
	journal={IEEE Transactions on Image Processing},
	volume={30},
	pages={3885--3896},
	year={2021},
	publisher={IEEE}
}

@inproceedings{lin2023event,
	title={Event-guided Frame Interpolation and Dynamic Range Expansion of Single Rolling Shutter Image},
	author={Lin, Guixu and Han, Jin and Cao, Mingdeng and Zhong, Zhihang and Zheng, Yinqiang},
	booktitle={Proceedings of the 31st ACM International Conference on Multimedia},
	pages={3078--3088},
	year={2023}
}

@inproceedings{zou2021learning,
	title={Learning to reconstruct high speed and high dynamic range videos from events},
	author={Zou, Yunhao and Zheng, Yinqiang and Takatani, Tsuyoshi and Fu, Ying},
	booktitle={Proceedings of the IEEE/CVF Conference on Computer Vision and Pattern Recognition},
	pages={2024--2033},
	year={2021}
}

@inproceedings{dai2016rolling,
	title={Rolling shutter camera relative pose: Generalized epipolar geometry},
	author={Dai, Yuchao and Li, Hongdong and Kneip, Laurent},
	booktitle={Proceedings of the IEEE conference on computer vision and pattern recognition},
	pages={4132--4140},
	year={2016}
}

@article{meingast2005geometric,
	title={Geometric models of rolling-shutter cameras},
	author={Meingast, Marci and Geyer, Christopher and Sastry, Shankar},
	journal={arXiv preprint cs/0503076},
	year={2005}
}

@article{liang2008analysis,
	title={Analysis and compensation of rolling shutter effect},
	author={Liang, Chia-Kai and Chang, Li-Wen and Chen, Homer H},
	journal={IEEE transactions on image processing},
	volume={17},
	number={8},
	pages={1323--1330},
	year={2008},
	publisher={IEEE}
}

@inproceedings{baker2010removing,
	title={Removing rolling shutter wobble},
	author={Baker, Simon and Bennett, Eric and Kang, Sing Bing and Szeliski, Richard},
	booktitle={2010 IEEE Computer Society Conference on Computer Vision and Pattern Recognition},
	pages={2392--2399},
	year={2010},
	organization={IEEE}
}

@inproceedings{naor2022combining,
	title={Combining internal and external constraints for unrolling shutter in videos},
	author={Naor, Eyal and Antebi, Itai and Bagon, Shai and Irani, Michal},
	booktitle={European Conference on Computer Vision},
	pages={119--134},
	year={2022},
	organization={Springer}
}

@inproceedings{zhou2022evunroll,
	title={Evunroll: Neuromorphic events based rolling shutter image correction},
	author={Zhou, Xinyu and Duan, Peiqi and Ma, Yi and Shi, Boxin},
	booktitle={Proceedings of the IEEE/CVF Conference on Computer Vision and Pattern Recognition},
	pages={17775--17784},
	year={2022}
}

@inproceedings{fan2021sunet,
	title={Sunet: symmetric undistortion network for rolling shutter correction},
	author={Fan, Bin and Dai, Yuchao and He, Mingyi},
	booktitle={Proceedings of the IEEE/CVF International Conference on Computer Vision},
	pages={4541--4550},
	year={2021}
}

@inproceedings{ji2023rethinking,
	title={Rethinking Video Frame Interpolation from Shutter Mode Induced Degradation},
	author={Ji, Xiang and Wang, Zhixiang and Zhong, Zhihang and Zheng, Yinqiang},
	booktitle={Proceedings of the IEEE/CVF International Conference on Computer Vision},
	pages={12259--12268},
	year={2023}
}

@inproceedings{nah2017deep,
	title={Deep multi-scale convolutional neural network for dynamic scene deblurring},
	author={Nah, Seungjun and Hyun Kim, Tae and Mu Lee, Kyoung},
	booktitle={Proceedings of the IEEE conference on computer vision and pattern recognition},
	pages={3883--3891},
	year={2017}
}

@inproceedings{huang2022real,
	title={Real-time intermediate flow estimation for video frame interpolation},
	author={Huang, Zhewei and Zhang, Tianyuan and Heng, Wen and Shi, Boxin and Zhou, Shuchang},
	booktitle={European Conference on Computer Vision},
	pages={624--642},
	year={2022},
	organization={Springer}
}

@inproceedings{weng2023event,
	title={Event-based blurry frame interpolation under blind exposure},
	author={Weng, Wenming and Zhang, Yueyi and Xiong, Zhiwei},
	booktitle={Proceedings of the IEEE/CVF Conference on Computer Vision and Pattern Recognition},
	pages={1588--1598},
	year={2023}
}

@inproceedings{gehrig2020video,
	title={Video to events: Recycling video datasets for event cameras},
	author={Gehrig, Daniel and Gehrig, Mathias and Hidalgo-Carri{\'o}, Javier and Scaramuzza, Davide},
	booktitle={Proceedings of the IEEE/CVF Conference on Computer Vision and Pattern Recognition},
	pages={3586--3595},
	year={2020}
}

@article{kingma2014adam,
	title={Adam: A method for stochastic optimization},
	author={Kingma, Diederik P and Ba, Jimmy},
	journal={arXiv preprint arXiv:1412.6980},
	year={2014}
}

@inproceedings{zhong2023blur,
	title={Blur interpolation transformer for real-world motion from blur},
	author={Zhong, Zhihang and Cao, Mingdeng and Ji, Xiang and Zheng, Yinqiang and Sato, Imari},
	booktitle={Proceedings of the IEEE/CVF Conference on Computer Vision and Pattern Recognition},
	pages={5713--5723},
	year={2023}
}

@inproceedings{zhong2022animation,
	title={Animation from blur: Multi-modal blur decomposition with motion guidance},
	author={Zhong, Zhihang and Sun, Xiao and Wu, Zhirong and Zheng, Yinqiang and Lin, Stephen and Sato, Imari},
	booktitle={European Conference on Computer Vision},
	pages={599--615},
	year={2022},
	organization={Springer}
}

@inproceedings{oh2022demfi,
	title={Demfi: deep joint deblurring and multi-frame interpolation with flow-guided attentive correlation and recursive boosting},
	author={Oh, Jihyong and Kim, Munchurl},
	booktitle={European Conference on Computer Vision},
	pages={198--215},
	year={2022},
	organization={Springer}
}

@inproceedings{kirillov2023segment,
	title={Segment anything},
	author={Kirillov, Alexander and Mintun, Eric and Ravi, Nikhila and Mao, Hanzi and Rolland, Chloe and Gustafson, Laura and Xiao, Tete and Whitehead, Spencer and Berg, Alexander C and Lo, Wan-Yen and others},
	booktitle={Proceedings of the IEEE/CVF international conference on computer vision},
	pages={4015--4026},
	year={2023}
}
\bibliographystyle{icml}

\newpage
\clearpage
\appendix
\renewcommand{\figurename}{Extended Figure}
\renewcommand{\tablename}{Extended Table}

\renewcommand\thefigure{\thesection.\arabic{figure}} 
\setcounter{figure}{0}   

\renewcommand\thetable{\thesection.\arabic{table}} 
\setcounter{table}{0}


\section{Implementation Details}
\label{implementation_details}

\begin{table}[!th]
	\centering
	\caption{Specifications of our coaxial imaging system. The deadtime between two adjacent high-speed frames of GS1 and GS2 is extremely short and can therefore be ignored.
	}
	\vspace{-2mm}
	\setlength\tabcolsep{4pt}	
	\resizebox{1\linewidth}{!}
	{
		\begin{tabular}{lcccc}
			\toprule
			\textbf{Device} & \textbf{RSGR} & \textbf{iRSGR} & \textbf{GS 1} &  \textbf{GS 2}\\
			\midrule
			{Exp. per Row} & $2\sim20$ ms  & $2\sim20$ ms & 2 ms    & 2 ms\\ 
			{Delay. per Row} & 37.5 $\mu$s  & 37.5 $\mu$s & 0 $\mu$s  & 0 $\mu$s\\ 
			{Exp. per Frame} & 20 ms  & 20 ms & 2 ms  & 2 ms\\
			{Deadtime} & 20 ms  & 20 ms  & 0 ms  & 0 ms\\ 
			{Frame rate} & 25 fps & 25 fps & 500 fps & 500 fps\\
			{Resolution} & $640\!\times\!480$  & $640\!\times\!480$ & $640\!\times\!480$  & $640\!\times\!480$ \\  
			\bottomrule
		\end{tabular}
	}
	\label{tab:imaging_system}
	\vspace{-4mm}
\end{table}

\begin{figure}[!ht]
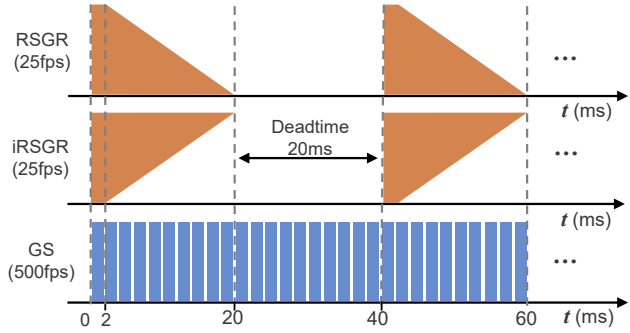

	\mfigure{1}{supp/sensor_align.pdf}
	\vspace{-5mm}
	\caption{
		Temporal alignment of sensors in our imaging system.
	}
	\label{fig:sensor_align}
	\vspace{-1mm}
\end{figure}

\begin{figure*}[!t]
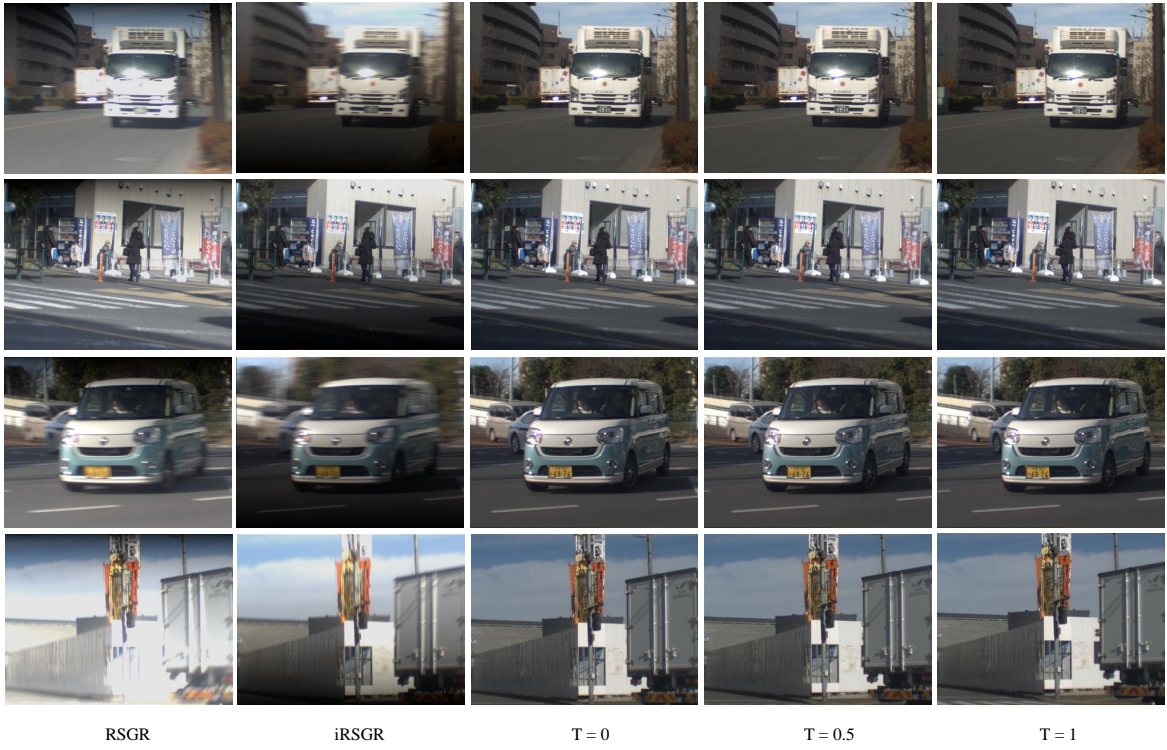

	\centering
	\mfigure{0.9}{supp/real_data.pdf}
	\caption{
		Data samples from our RSGR-HDR. RSGR view is exposed downwards while iRSGR scans upwards. We present three groundtruth frames which are temporally located at $T=0,0.5,1$. Here, the time instants are defined as normalized timestamps in $[0,1]$ over the latent video duration.
	}
	\label{fig:real_data}
\end{figure*}

\begin{figure*}[!t]
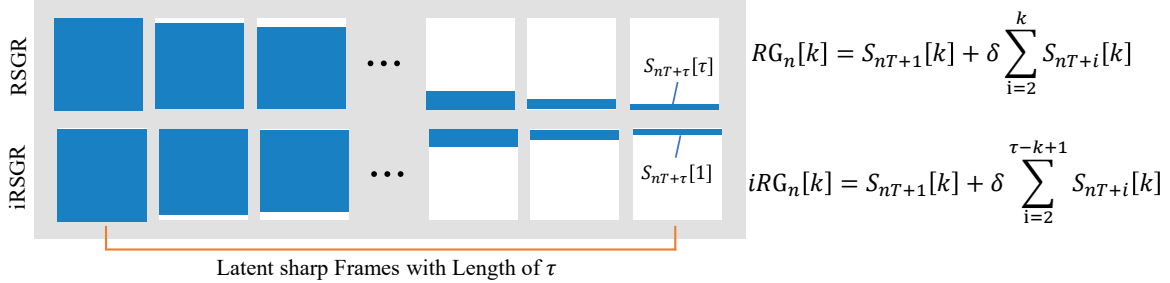

	\centering
	\mfigure{0.9}{supp/synthetic_process.pdf}
	\caption{
		Synthetic method. The notation $RG[k]$ denotes extracting the k-th row from frame $RG$. $n$ is the index of RSGR or iRSGR view. $T$ is the number of latent frames that correspond to exposure and deadtime. $RG$ denotes the RSGR images with scanning direction from top to bottom while $iRG$ has the inverted exposure direction. $\delta$ is the ratio between constant delay among rows and duration of the first exposed row.
	}
	\label{fig:synthetic_process}
	\vspace{-3mm}
\end{figure*}

\subsection{Construction of Imaging system}
\label{sec:system_details}
To construct a realistic dataset that contains perfectly aligned RSGR, iRSGR, and high-speed HDR video streams, we begin by fixing the RSGR camera and manually tuning the pose of the remaining cameras. This is guided by inspecting checkerboard residuals. However, achieving sub-pixel alignment for all four cameras across every axis proves extremely challenging due to the intricate nature of the multi-sensor rig. As a result, we eventually lock all components of the system in place and perform the alignment computationally. Specifically, we follow strategy in~\cite{rim2020real} and estimate three homographies under a closed-loop constraint to synchronize the viewpoints of the four sensors. The detailed specifications and acquisition parameters for each device are listed in~\tabref{imaging_system}.

The GS camera used in our system—BTRAN CS-700C equipped with a SONY IMX426 sensor—exhibits very low noise. Two factors mainly contribute to this. First, the IMX426 increases its effective pixel size to $9\mu m \times 9\mu m$  by combining four $4.5\mu m \times 4.5\mu m$ subpixels at the circuit level, resulting in high sensitivity and reduced read noise. Second, the camera is actively cooled to $0^\circ\mathrm{C}$, which further suppresses thermally induced noise.

To ensure that the system remains feasible in practice, we pair the imaging device with a FUJINON HF12.5HA-1S lens (focal length 12.5 mm, supporting 2/3-inch formats). Given the effective pixel pitch of $9\mu m$, the active sensing region of $640 \times 480$ corresponds to $5.7mm \times 4.3mm$. Because the relay lens operates at a $1:1$ magnification ratio, the horizontal and vertical fields of view become $26.1^\circ$ and $19.5^\circ$, respectively. At a distance of $10m$, this equates to a view coverage of $4.6m$ horizontally and $3.4m$ vertically, making the setup suitable for real-world capture. Although lens distortion may interact with rolling-shutter effects, selecting a high-grade FUJINON lens helps minimize such artifacts.

We follow the method proposed in~\cite{lin2023event,zou2021learning} to merge the two GS frames into HDR image. One of the high-speed sensors is fitted with a Thorlabs ND513B neutral-density filter to attenuate incoming light. ND filters reduce irradiance uniformly in both spatial and spectral domains. This enables us to record two GS frames with different illumination levels without altering exposure duration, which is often difficult to control reliably when capturing at high frame rates. The two complementary exposures provide the necessary variation for synthesizing HDR results.

\subsection{Experimental Data and Training Details}
\label{sec:data_training}
\subsubsection{Real Dataset}
The comprehensive methodology for collecting our real data is detailed in~\secref{data_capture}. Utilizing the capabilities of our imaging system, we collected the RSGR-HDR dataset, which comprises 90 urban scenes. Each scene includes $70 \times 2$ degraded frames and 1,400 aligned high-speed sharp frames. As shown in~\tabref{imaging_system} and \figref{sensor_align}, each RSGR–iRSGR pair is acquired over a 40 ms cycle, consisting of a 20 ms exposure followed by a 20 ms dead time. Meanwhile, the two GS cameras operate at 500 fps with a 2 ms exposure and no dead time, producing 20 synchronized GS pairs per cycle. Ten pairs fall within the RSGR–iRSGR exposure window and are fused into 10 temporally aligned HDR frames, while the remaining ten fall within the dead time. Therefore, in RSGR-HDR, each RSGR–iRSGR pair corresponds to 10 HDR GT frames. Representative RSGR-HDR samples are illustrated in~\figref{real_data}. The visual discrepancies between the two views are distinctly noticeable. The mutually inverted scanning manner allows the exposure compensation, which provides supplemental details and cues for robust HDR reconstruction. We further divide them into 54, 12 and 24 sequences as our training, validating, and testing set, respectively.

\begin{table*}[!thp]
	\centering
	\caption{Quantitative comparisons on GOPRO-Dual. We present experimental results on recovering a single latent HDR frame (HDR reconstruction) and an HDR sequence of length $3$, $5$, and $9$ (HDR photosequencing), both extracted within the exposure duration of the input frame. Performance is evaluated using mean PSNR/SSIM/LPIPS/TMQI. Dual reversed RSGR inputs are denoted as `RG-iRG'. `n$\cdot$RG' is the setting using $n$ neighboring RSGR frames and `RG-E' combines RSGR with event streams. Note that, LAN takes 5 consecutive LDR frames with alternate long and short exposures (denoted as '5$\cdot$LS').} 
	\label{tab:synthetic_comparison_full}
	\resizebox{\linewidth}{!}
	{
		\begin{tabular}{rccccc}
			\toprule
			\multirow{2}{*}{Method}& \multirow{2}{*}{Input} & \multirow{2}{*}{HDR Reconstruction} &\multicolumn{3}{c}{HDR Photosequencing}  \\ \cmidrule(lr){4-6}
			
			& & &   $\times3$  &$\times5$     & $\times9$  \\ \midrule
			
			DSUN ~\cite{liu2020deep}& \multirow{8}{*}{\emph{2}$\cdot$\emph{RG}} & 19.92 / 0.7850 / 0.1347 / 0.8691 & -- / -- / --  & -- / -- / -- & -- / -- / --   \\ 
			
			JAMNet ~\cite{fan2023joint}& & 25.16 / 0.8797 / 0.0940 / 0.9186 & -- / -- / --  & -- / -- / -- & -- / -- / --   \\ 
			
			RSSR ~\cite{fan2021inverting}&  & 12.21 / 0.6535 / 0.2394 / 0.8164 & 12.86 / 0.6860 / 0.2054 / 0.8033 & 12.94 / 0.6918 / 0.2079 / 0.8052 & 13.02 / 0.6908 / 0.1956 / 0.8060  \\ 
			
			CVR ~\cite{fan2022context}&   & 13.29 / 0.6886 / 0.2291 / 0.8211 &  13.08 / 0.6768 / 0.2121 / 0.8069 & 13.14 / 0.6797 / 0.2164 / 0.8005 &  12.98 / 0.6699 / 0.2062 / 0.7978 \\
			
			RIFE~\cite{huang2022real} & &  20.97 / 0.8327 / 0.1646 / 0.9412&  20.91 / 0.8319 / 0.1756 / 0.8868 & 20.94 / 0.8330 / 0.1728 / 0.8896 & 20.96 / 0.8335 / 0.1710 / 0.8910 \\
			
			AfB~\cite{zhong2022animation} & &  20.64 / 0.7021 / 0.3080 / 0.9354 & 20.85 / 0.7251 / 0.3165 / 0.8518 & 20.84 / 0.7212 / 0.3118 / 0.8540 & 20.84 / 0.7204 / 0.3092 / 0.8553 \\
			IFED~\cite{zhong2022bringing}&  & 23.96 / 0.8447 / 0.0949 / 0.9431 &  24.27 / 0.8404 / 0.1036 / 0.8943 &  24.24 / 0.8424 / 0.1011 / 0.8974 & 24.24 / 0.8434 / 0.1000 / 0.8989 \\
			
			PMBNet~\cite{ji2023rethinking}&  & 25.42 / 0.9009 / 0.1061 / 0.9252 &  24.94 / 0.8870 / 0.1198 / 0.9207 & 25.07 / 0.8907 / 0.1167 / 0.9191 & 25.11 / 0.8923 / 0.1160 / 0.9180  \\ \midrule
			
			BiT~\cite{zhong2023blur} & \emph{3}$\cdot$\emph{RG} &  23.96 / 0.8673 / 0.1034 / 0.9392 &  24.48 / 0.9011 / 0.0837 / 0.9302  & 24.40 / 0.8952 / 0.0863 / 0.9287  & 24.35 / 0.8921 / 0.0882 / 0.9273 \\ \midrule
			
			ExpandNet~\cite{marnerides2018expandnet} & \multirow{2}{*}{\emph{1}$\cdot$\emph{RG}} & 17.59 / 0.7410 / 0.2702 / 0.8274 & -- / -- / --  & -- / -- / -- & -- / -- / --    \\
			
			RSS-T~\cite{ji2023single} &   & 23.00 / 0.9229 / 0.0756 / 0.9316 &  -- / -- / --   & -- / -- / -- & -- / -- / --  \\ \midrule
			
			DeMFI~\cite{oh2022demfi} & \emph{4}$\cdot$\emph{RG} &  21.06 / 0.8520 / 0.1324 / 0.9203 &  20.75 / 0.8288 / 0.1847 / 0.8997  & 20.84 / 0.8348 / 0.1763 / 0.8940 & 20.59 / 0.8191 / 0.2014 / 0.8910 \\ \midrule

			NGS~\cite{wang2022neural}& \emph{5}$\cdot$\emph{RG} & 31.04 / 0.9645 / 0.0343 / 0.9457 & -- / -- / --  & -- / -- / -- & -- / -- / --    \\ \midrule
			
			LAN~\cite{chung2023lan} &  \multirow{2}{*}{\emph{5}$\cdot$\emph{LS}}  & 34.38 / 0.9416 / 0.0589 / 0.7003 &  -- / -- / --   & -- / -- / -- & -- / -- / --  \\ 
			HDRFlow~\cite{xu2024hdrflow}& & 21.17 / 0.4041 / 0.4663 / 0.6520 & -- / -- / --  & -- / -- / -- & -- / -- / --   \\  \midrule
			
			EBFI~\cite{weng2023event}& \multirow{4}{*}{\emph{RG}-\emph{E}} & 20.82 / 0.8864 / 0.1034 / 0.9354 & 20.73 / 0.8824 / 0.1040 / 0.9230  & 20.75 / 0.8837 / 0.1036 / 0.9232 & 20.77 / 0.8844 / 0.1034 / 0.9233   \\
			
			EvUnroll~\cite{zhou2022evunroll}&   & 29.27 / 0.9372 / 0.0496 / 0.9446 &  29.07 / 0.9321 / 0.0522 / 0.9348   & 29.13 / 0.9336 / 0.0513 / 0.9350 & 29.17 / 0.9345 / 0.0509 / 0.9351 \\
			
			EFI~\cite{lin2023event}&   & 20.45 / 0.8905 / 0.1005 / 0.9427 & 20.38 / 0.8861 / 0.1047 / 0.9371   & 20.40 / 0.8872 / 0.1032 / 0.9373 & 20.41 / 0.8874 / 0.1027 / 0.9373  \\ 
			UniINR~\cite{lu2024uniinr}&   & 24.21 / 0.9124 / 0.1090 / 0.9335 & 23.31 / 0.8495 / 0.1731 / 0.9120   & 23.30 / 0.8468 / 0.1763 / 0.9114 & 23.31 / 0.8448 / 0.1794 / 0.9111  \\ \midrule
			
			RIFE$_m$~\cite{huang2022real} & \multirow{3}{*}{\emph{RG}-\emph{iRG}} & 32.39 / 0.9591 / 0.0659 / 0.9544 & 33.56 / 0.9475 / 0.0710 / 0.9399 & 32.89 / 0.9502 / 0.0712 / 0.9407 & 32.49 / 0.9516 / 0.0712 / 0.9412   \\
			
			IFED$_m$~\cite{zhong2022bringing}&  & 32.34 / 0.9625 / \textbf{0.0291} / 0.9542 &  32.64 / 0.9437 / \textbf{0.0341} / 0.9404 & 32.25 / 0.9486 / \textbf{0.0331} / 0.9420  & 32.05 / 0.9509 / \textbf{0.0327} / 0.9427  \\
			
			BiT$_m$~\cite{zhong2023blur}&  & 34.17 / 0.9733 / 0.0400 / 0.9538 & 33.68 / 0.9569 / 0.0477 / 0.9428 & 33.58 / 0.9614 / 0.0458 / 0.9439  & 33.53 / 0.9637 / 0.0449 / 0.9444  \\
			
			Ours &  & \textbf{35.88} / \textbf{0.9795} / 0.0293 / \textbf{0.9546} & \textbf{36.41} / \textbf{0.9708} / 0.0378 / \textbf{0.9471}  &  \textbf{35.88} / \textbf{0.9729} / 0.0369 / \textbf{0.9473} & \textbf{35.55} / \textbf{0.9738} / 0.0364 / \textbf{0.9474} \\ \bottomrule
		\end{tabular}
	}
\end{table*}

\begin{figure}[!t]
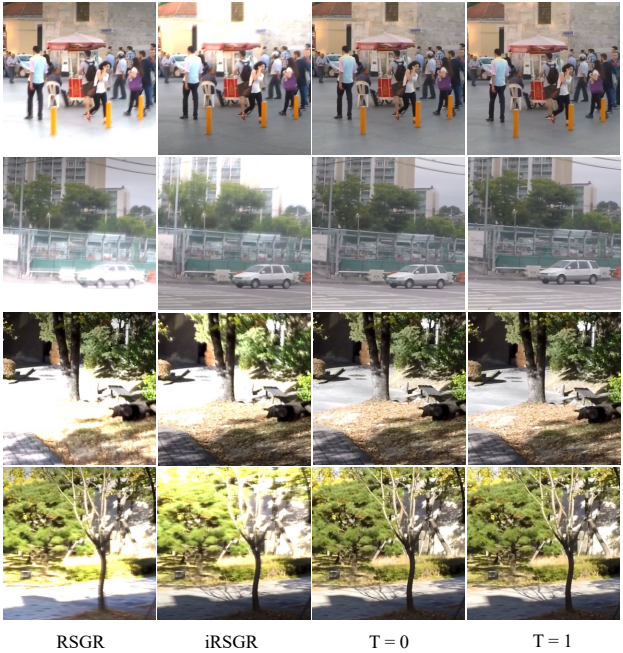

	\vspace{2mm}
	\centering
	\mfigure{1}{supp/synthetic_data.pdf}
	\vspace{-2mm}
	\caption{
		Data samples from our synthetic GOPRO-Dual dataset. For each dual reversed RSGR pair, we present two groundtruth HDR frames here which are temporally located at $T=0,1$.
	}
	\label{fig:synthetic_data}
\end{figure}

\begin{figure*}[!t]
	\centering
	\mfigure{0.9}{exp/sota_comparison_goprovfi.pdf}
	\caption{
		Qualitative comparison on GOPRO-Dual. Our model outperforms the existing approaches with different settings in details recovery and dynamic range expansion.
	}
	\label{fig:sota_comparison_goprovfi_full}
\end{figure*}

\subsubsection{Synthetic Dataset}
\label{sec:synthesizing_process}

To strengthen our study of how dual reversed RSGR setting aids HDR photosequencing, we also developed a synthetic dataset called GOPRO-Dual, adhering to the methodologies specified in \cite{ji2023single,wang2022neural}. While synthetic data cannot fully capture the complexity and diversity of real-world scenes, it is used here only for supplementary and independent validation. Our main training and evaluation are conducted on the real-world RSGR-HDR dataset. Meanwhile, experiments on synthetic data further verify our findings and provide a meaningful reference for specific analyses (e.g., ensuring fair comparison across methods). This dataset draws from the GOPRO data\cite{nah2017deep}, which encompasses 33 videos with a resolution of $1280 \times 720$. Each video features 1200 consecutive frames shot at 240fps. To render the dataset more lifelike, we significantly enhanced the framerate of the GOPRO videos, increasing them by a factor of $\times 64$ through a widely-used video interpolation algorithm~\cite{huang2022real}.

~\figref{synthetic_process} illustrates our synthesizing method. We formulate the imaging process of RSGR as weighted sum of patches extracted from corresponding latent sharp frames within exposure time $\tau$~\cite{ji2023single,wang2022neural}. RSGR view is synthesized with scanning direction from top to bottom, while iRSGR exposes frames upwards. We strictly constrain the generation process as shown in  \figref{synthetic_process}, the two reversed RSGR views are aligned in frame level to ensure they observe the identical visual content of scenes. Specifically, the original GOPRO frames are centrally cropped as $512\times512$ and we set $T = \tau = 512$, which leads to 512 sharp frames corresponding to an RSGR or iRSGR frame. For enabling supervised learning, we uniformly sampled 9 frames out of them to take as ground truth.~\figref{synthetic_data} exhibits some examples of synthesized data.

In the experiments, we compared our method with some competitive settings: EFI~\cite{lin2023event}, EvUnroll~\cite{zhou2022evunroll} and EBFI~\cite{weng2023event} assisted by \emph{event camera}, LAN~\cite{chung2023lan} using \emph{alternating exposure} within an input sequence. However, our collected dataset: RSGR-HDR and GOPRO-Dual, exclude the event or multi-exposure counterpart. Thus, we synthesized the event stream from the high frame-rate videos using an event simulator~\cite{gehrig2020video}. The alternating exposure sequence could be generated based on the HDR frames through mapping strategy provided in~\cite{chen2021hdr}.

\subsubsection{Training Details}

During learning process, each training sample consists of RSGR-iRSGR pairs and a HDR video clip with frame length of $9$. We adopt Adam optimizer~\cite{kingma2014adam} with initial learning rate of $10^{-4}$ combined with a cosine annealing scheduler. The total training epoch is set as $800$. To augment data, training samples are randomly cropped to $256 \times 256$, then flipped vertically and horizontally. The whole training process is performed on two GPUs of NVIDIA GeForce RTX 5090 with batch size of 8. Conventionally, we assess the performance of our models using established metrics: PSNR, SSIM, and LPIPS. 


\section{Additional Results}
\label{results}

\subsection{Experimental Results on Synthetic Data}
In~\secref{experiments_synthesic}, we compare our model with some selected SOTAs on synthetic data. The complete quantitative and qualitative results are shown in \tabref{synthetic_comparison_full} and \figref{sota_comparison_goprovfi_full}, respectively.

\begin{figure}[!tp]
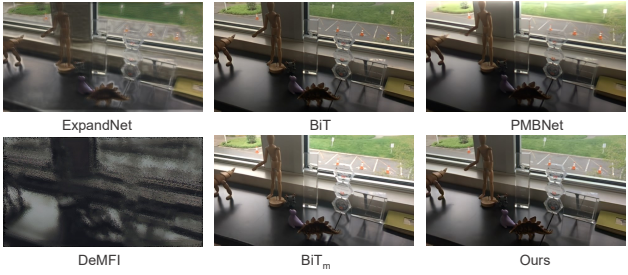

	\centering
	\mfigure{1}{exp/thirdparty_test.pdf}
	\vspace{-2mm}
	\caption{
		\textbf{Qualitative comparison} with representative SOTAs on Adobe-Dual testset.
	}
	\label{fig:thirdparty_test}
	\vspace{-2mm}
\end{figure}

\begin{table}[!tp]
	\centering
	\caption{\textbf{Quantitative comparison} with representative SOTAs on Adobe-Dual testset.}
	\label{tab:thirdparty_test}
	\resizebox{1\linewidth}{!}
	{
		\begin{tabular}{rccccccc}
			\toprule
			\multirow{2}{*}{Method}& \multirow{2}{*}{Input} & \multicolumn{3}{c}{$\times$1} & \multicolumn{3}{c}{$\times$9} \\ \cmidrule(lr){3-5}\cmidrule(lr){6-8}
			
			& & PSNR     & SSIM     & LPIPS    & PSNR    & SSIM    & LPIPS    \\ \midrule

			ExpandNet~\cite{marnerides2018expandnet} &  \emph{1}$\cdot$\emph{RG}  & 15.75 & 0.7220 & 0.1877  & -- & -- & --    \\  \midrule

			PMBNet~\cite{ji2023rethinking} & \emph{2}$\cdot$\emph{RG} & 16.79  & 0.7631 & 0.1450 & 17.36  & 0.7550 & 0.1632   \\  \midrule
			
			BiT~\cite{zhong2023blur}& \emph{3}$\cdot$\emph{RG}  & 18.79 &  0.8187 & 0.1291 & 18.29 & 0.7676  & 0.1481   \\ \midrule 			
			
			DeMFI~\cite{oh2022demfi} &  \emph{4}$\cdot$\emph{RG}  & 13.94  & 0.4089 & 0.4709 & 13.29 & 0.3717 &  0.5231    \\ \midrule
			
			BiT$_m$~\cite{zhong2023blur} &  \multirow{2}{*}{\emph{RG}$\cdot$\emph{iRG}}  & 22.48 & 0.8802 & 0.0954   & 21.18 & 0.8270 & 0.1169    \\
			Ours\quad\,\,\,\,\, &  & 23.54 & 0.8950  & 0.0775  & 22.25 & 0.8463 & 0.0964    \\ \bottomrule
		\end{tabular}
	}
	\vspace{-2mm}
\end{table}

\subsection{Third-party Evaluation}

To further evaluate the generalization ability of our framework, we additionally construct two third-party testsets, Adobe-Dual and LiU-HDRv-Dual, based on Adobe240~\cite{su2017deep} and the LiU HDRv~\cite{kronander2014unified}, respectively. Adobe-Dual contains diverse scene contents, motion patterns, illumination conditions, and synthesis parameters that differ from those of our training data, while LiU-HDRv-Dual is introduced to evaluate the performance of our method on conventional HDR scenes with pronounced illumination variations, dark regions, and saturated highlights. Following the same dual reversed RSGR imaging pipeline described in~\secref{synthesizing_process}, we synthesize the corresponding RSGR-iRSGR observations and HDR targets for both testsets. In all experiments, the pretrained model is directly evaluated without any finetuning or domain adaptation.

\begin{figure*}[!tp]
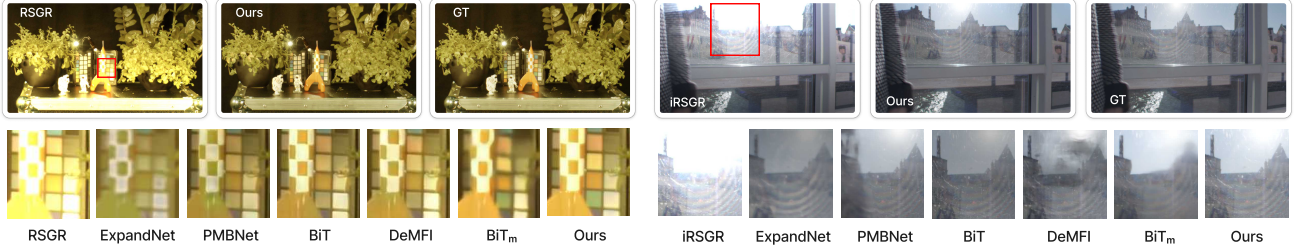

	\centering
	\mfigure{1}{exp/thirdparty_test_liuhdrv.pdf}
	\vspace{-4mm}
	\caption{
		\textbf{Qualitative comparison} with representative SOTAs on LiU-HDRv-Dual testset. Best viewed with zoom.
	}
	\label{fig:thirdparty_test2}
	\vspace{-2mm}
\end{figure*}

\tabref{thirdparty_test} and~\figref{thirdparty_test} present the quantitative and visual comparisons on Adobe-Dual, demonstrating the robustness of our method to unseen scenes, motions, synthesis parameters, and illumination conditions. Meanwhile, the results on LiU-HDRv-Dual in~\tabref{thirdparty_test2} and~\figref{thirdparty_test2} further verify its generalization to conventional HDR scenes with challenging underexposed and overexposed regions. On both testsets, our method achieves the best overall performance and consistently reconstructs fine temporal details with a wide dynamic range.

\begin{table}[!tp]
	\centering
	\caption{\textbf{Quantitative comparison} with representative SOTAs on LiU-HDRv-Dual testset.}
	\label{tab:thirdparty_test2}
	\resizebox{1\linewidth}{!}
	{
		\begin{tabular}{rccccccc}
			\toprule
			\multirow{2}{*}{Method}& \multirow{2}{*}{Input} & \multicolumn{3}{c}{$\times$1} & \multicolumn{3}{c}{$\times$9} \\ \cmidrule(lr){3-5}\cmidrule(lr){6-8}
			
			& & PSNR     & SSIM     & LPIPS    & PSNR    & SSIM    & LPIPS    \\ \midrule

			ExpandNet~\cite{marnerides2018expandnet} &  \emph{1}$\cdot$\emph{RG}  & 17.45 & 0.7274 & 0.2987  & -- & -- & --    \\  \midrule

			PMBNet~\cite{ji2023rethinking} & \emph{2}$\cdot$\emph{RG} & 19.41  & 0.7753 & 0.2374 & 19.39  & 0.7753 & 0.2544   \\  \midrule
			
			BiT~\cite{zhong2023blur}& \emph{3}$\cdot$\emph{RG}  & 19.94 &  0.7687 & 0.2460 & 20.36 & 0.7964  & 0.1861   \\ \midrule 			
			
			DeMFI~\cite{oh2022demfi} &  \emph{4}$\cdot$\emph{RG}  & 17.42  & 0.7257 & 0.2582 & 17.66 & 0.7254 &  0.2969    \\ \midrule
			
			BiT$_m$~\cite{zhong2023blur} &  \multirow{2}{*}{\emph{RG}$\cdot$\emph{iRG}}  & 22.14 & 0.8323 & 0.2165   & 22.80 & 0.8282 & 0.2111    \\
			Ours\quad\,\,\,\,\, &  & 24.24 & 0.8604  & 0.1538  & 24.72 & 0.8567 & 0.1512    \\ \bottomrule
		\end{tabular}
	}
	\vspace{-2mm}
\end{table}

These results suggest that the superiority of our framework does not mainly originate from overfitting to a specific dataset distribution, but rather from the intrinsic row-wise complementary property of dual reversed RSGR inputs (BiT \vs BiT$_m$) together with the proposed row-adaptive alignment and correlation-guided hallucination modules (Ours \vs BiT$_m$).

\begin{figure*}[!tbh]
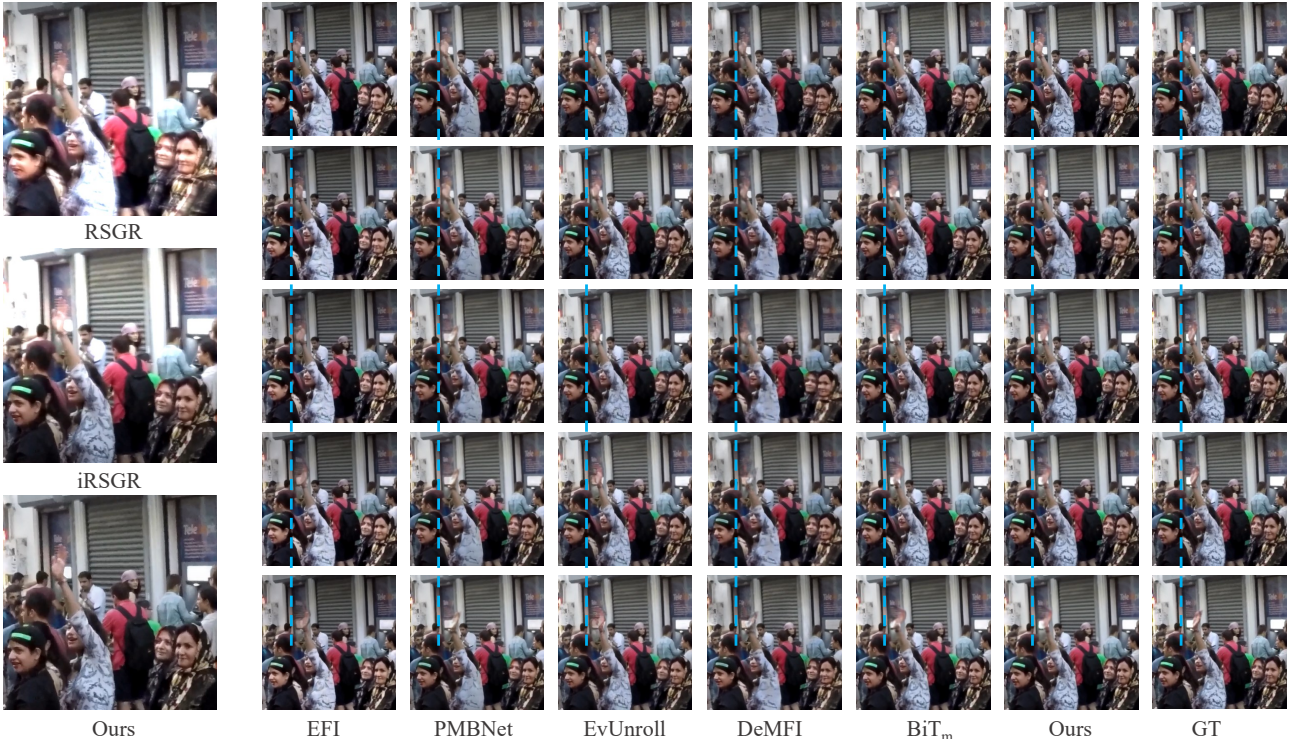

	\centering
	{
		\mfigure{1}{supp/video_reconstruction_gopro_v2.pdf}
	}  
	
	\vspace{-2mm}
	\caption{
		Qualitative comparison on HDR photosequencing of GOPRO-Dual. We present the multiple intermediate frames (5 out of 9) at different time generated by different models. Each row denotes reconstructed latent frames temporally located at $t=[0,2/8,4/8,6/8,1]$. Best viewed with zoom.
	}
	\label{fig:video_reconstruction}
	\vspace{-2mm}
\end{figure*}

\subsection{Video reconstruction results}
\label{sec:video_recon}
\begin{figure*}[!tbh]
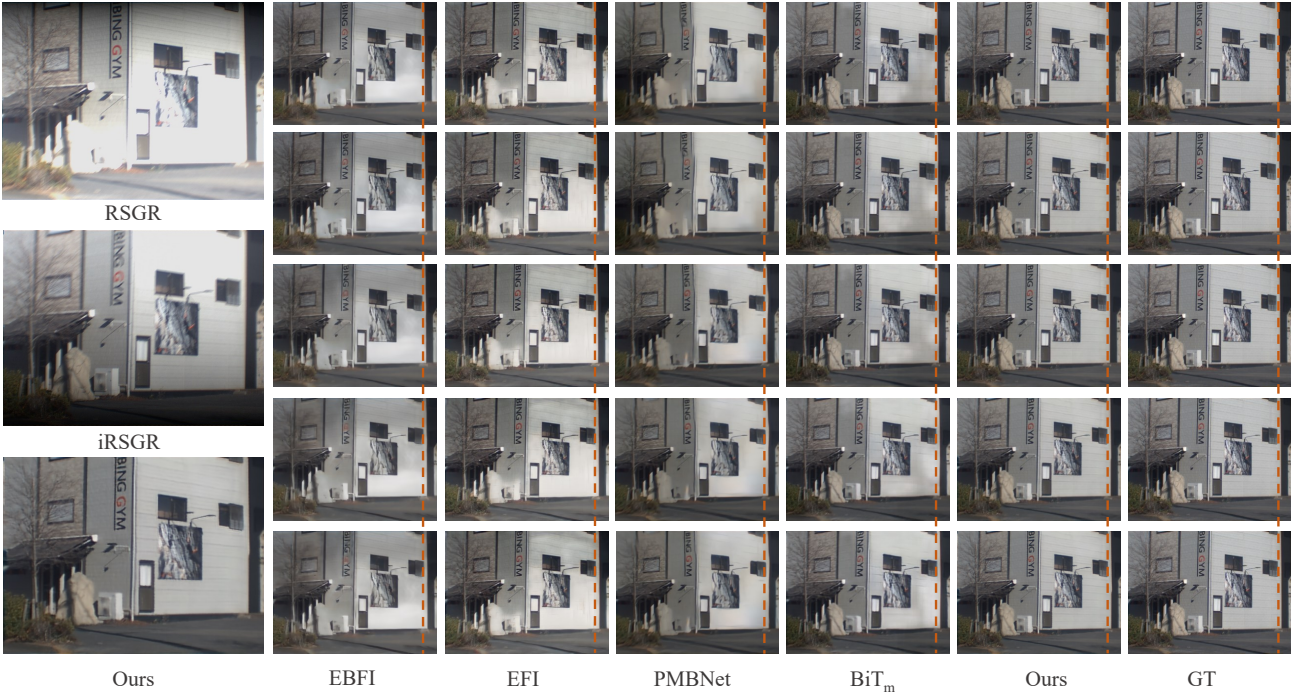

	\centering
	{
		\mfigure{1}{supp/video_reconstruction_rsgrhdr.pdf}
	}  
	
	\caption{
		Qualitative comparison on HDR photosequencing of RSGR-HDR. We present the multiple intermediate frames (5 out of 9) at different time generated by different models. Each row denotes reconstructed latent frames temporally located at $t=[0,2/8,4/8,6/8,1]$. Best viewed with zoom.
	}
	\label{fig:video_reconstruction_rsgrhdr}
\end{figure*}

To better substantiate the temporal consistency of our model in HDR photosequencing, we apply our model and competitive SOTAs to generate multiple consecutive latent HDR frames, whose visual results on GOPRO-Dual are given in ~\figref{video_reconstruction}. In terms of motion prediction, EFI fails to distinguish the movement of waving hand throughout all latent frames. The other approaches, such as PMBNet, EvUnroll and DeMFI, are able to disambiguate temporal direction but also introduce significant artifacts and distortions. Compared with BiTm, our results are clearer and closer to ground truths. Consistent observation can be found in the real data presented in~\figref{video_reconstruction_rsgrhdr}.

\subsection{Additional Qualitative Results}
\label{sec:more_results}
We present additional qualitative comparisons on RSGR-HDR and GOPRO-Dual in ~\figref{sota_comparison_supp} -- ~\figref{sota_comparison6}.


\begin{figure*}[!tb]
	\centering
	\mfigure{0.9}{supp/sota_comparison_supp.pdf}
	\caption{
		Qualitative comparisons on RSGR-HDR. Supplementary qualitative results corresponding to~\figref{sota_comparison} in the main manuscript.
	}
	\label{fig:sota_comparison_supp}
\end{figure*}


\begin{figure*}[!tb]
	\centering
	\mfigure{0.9}{supp/sota_comparison1.pdf}
	\caption{
		Additional qualitative comparisons on RSGR-HDR.
	}
	\label{fig:sota_comparison1}
\end{figure*}


\begin{figure*}[!tb]
	\centering
	\mfigure{0.9}{supp/sota_comparison2.pdf}
	\caption{
		Additional qualitative comparisons on RSGR-HDR.
	}
	\label{fig:sota_comparison2}
\end{figure*}


\begin{figure*}[!tb]
	\centering
	\mfigure{0.9}{supp/sota_comparison3.pdf}
	\caption{
		Additional qualitative comparisons on RSGR-HDR.
	}
	\label{fig:sota_comparison3}
\end{figure*}


\begin{figure*}[!tb]
	\centering
	\mfigure{0.9}{supp/sota_comparison4.pdf}
	\caption{
		Additional qualitative comparisons on GOPRO-Dual.
	}
	\label{fig:sota_comparison4}
\end{figure*}


\begin{figure*}[!tb]
	\centering
	\mfigure{0.9}{supp/sota_comparison5.pdf}
	\caption{
		Additional qualitative comparisons on GOPRO-Dual.
	}
	\label{fig:sota_comparison5}
\end{figure*}


\begin{figure*}[!tb]
	\centering
	\mfigure{0.9}{supp/sota_comparison6.pdf}
	\caption{
		Additional qualitative comparisons on GOPRO-Dual.
	}
	\label{fig:sota_comparison6}
\end{figure*}


\end{document}